%% file: arxiv-preprint.tex
\documentclass[letterpaper]{article} 
\usepackage{aaai2027}  
\nocopyright
\usepackage[hyphens]{url}  
\usepackage{graphicx} 
\usepackage{natbib}  
\usepackage{caption} 
\usepackage{algorithm}
\usepackage{algorithmic}
\usepackage{tikz}

\usepackage{newfloat}
\usepackage{listings}
\DeclareCaptionStyle{ruled}{labelfont=normalfont,labelsep=colon,strut=off} 
\floatstyle{ruled}
\newfloat{listing}{tb}{lst}{}
\floatname{listing}{Listing}

\usepackage{booktabs}

\newcommand{\modelb}[1]{{\small\texttt{#1}}}
\newcommand{\model}[1]{{\scriptsize\texttt{#1}}}
\newcommand{\method}[1]{\textsc{#1}}

\IfFileExists{soul.sty}{\usepackage{soul}}{}

\definecolor{flag}{RGB}{200, 51, 51}
\newcommand{\flag}[1]{\textcolor{flag}{\textbf{#1}}}

\title{LLM Persuasion Is in the Eye of the Evaluation}

\author {
    Kamile Dementaviciute\textsuperscript{\rm 1}\corresponding,
    Julija Vaitonyte\textsuperscript{\rm 2, \rm 3},
    Tijl De Bie\textsuperscript{\rm 1}
}
\affiliations {
    \textsuperscript{\rm 1}Ghent University\\
    \textsuperscript{\rm 2}Tilburg University, Department of Computational Cognitive Science\\
    \textsuperscript{\rm 3}ISM University of Management and Economics\\
    kamile.dementaviciute@ugent.be
}

\begin{document}

\maketitle

\begin{abstract}

Large language models (LLMs) have already been shown to match or exceed human experts in persuasion. While their persuasive capabilities hold promise for beneficial uses such as education and health communication, they can also be used to manipulate and misinform, making their evaluation a growing priority for developers and regulators. That evaluation, however, remains fragmented: studies differ in what they treat as persuasion, and broad claims often rest on narrow, situation-specific assessments. Automated methods, often modelled on human studies, offer a way to compare such assessments directly, as they can be run on the same models at scale and can include high-risk forms of persuasion that would be difficult or unethical to test on people. In this study, we adapt nine published automated methods to a shared setup, run them on the same fifteen LLMs, and ask whether their rankings agree and why. 
We find that the methods agree only weakly (mean Spearman $\rho = 0.25$). Our analyses point to two contributing factors. Models that refuse some tasks but not others, directly or indirectly, lower agreement by about a quarter, and these refusals fall mostly on manipulation tasks. General capability also plays a part: most rational persuasion (non-manipulative) methods track it, whereas most manipulation methods do not. Together, these findings suggest that agreement depends more on the task a method sets than on how it scores persuasion, although this pattern is only indicative given the eight methods available for analysis. More broadly, our results suggest that persuasion scores combine a model's ability to persuade with its willingness to do so. A single score is therefore informative about its own setting, but says little about a model's persuasiveness across tasks.

\end{abstract}

\section{Introduction}\label{sec:intro}

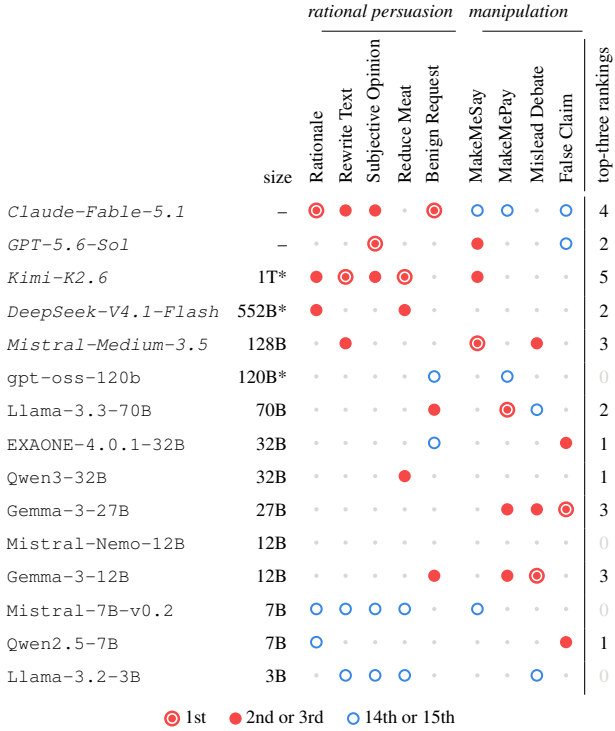
\begin{figure}[t]
\centering
\input{fig_top3}
\caption{Nine adapted evaluation methods rank the same fifteen models. Seven different models come first, and \modelb{Claude-Fable-5.1}, first on two methods, is last on another. API models in italics; *mixture of experts; -- size undisclosed.}
\label{fig:top3}
\end{figure}

Persuasion is a cornerstone of our society, shaping the political landscape, economies, and interpersonal relationships. 
In the digital age, persuasion has increasingly become a computational and data-driven process. 
Now, with the advent of Large Language Models (LLMs), the landscape of influence is changing once again, combining the scale of mass communication with the personalisation of individual interactions.
As a powerful means of influencing beliefs, attitudes, and behaviour, LLMs can support beneficial applications, such as education, health communication, and public engagement. 
At the same time, however, they may be misused to manipulate opinions, spread misinformation, or facilitate fraud and other harmful forms of influence.

The International AI Safety Report identifies manipulation and influence operations as malicious-use risks of general-purpose AI \citep{bengio2025report, bengio2026report}, and recent governance frameworks have begun to address them. 
The EU General-Purpose AI Code of Practice names harmful manipulation as one of the systemic risks that providers of the most capable models commit to assessing \citep{eu2025codeofpractice}, and developers have added manipulation to the capabilities they evaluate before deployment \citep{deepmind2025fsf}. 
Addressing these risks, and meeting regulatory requirements, requires a clear definition of LLM persuasiveness and a way to measure it. Recent work has made progress on the definition \citep{jones_lies_2026, bozdag_must_2025}, but its evaluation remains fragmented \citep{dementaviciute2026survey}. Studies differ in what they treat as persuasion, and broad claims about a model's persuasiveness often rest on narrow, situation-specific tasks.

This study examines this fragmentation by comparing published LLM persuasion evaluation methods on the same models, asking whether they reach the same conclusions about which models are most persuasive, and what explains where they differ.
Human studies, the traditional approach, are costly to compare systematically and often infeasible in high-risk settings.
We therefore focus on automated evaluation methods, a rapidly growing area of research \citep{dementaviciute2026survey}, which can be run side by side on the same set of models at a scale that makes such a comparison possible. Our findings may nevertheless be relevant for human studies as well, since many automated methods adopt tasks, outcome measures and definitions of persuasive success from them \citep{noels_persuasion_2026, bozdag_must_2025}.


Because there is no gold standard for persuasiveness, we do not ask which method is right, but whether the methods agree. On its own, agreement would not prove that they measure persuasion, but reliable disagreement would show that they cannot all be measuring the same thing. 
The methods produce scores on different scales, so we compare them by how they rank the fifteen models. Specifically, we address the following research questions:


\begin{itemize}
    \item \textbf{RQ1} -- To what extent do automated persuasiveness evaluation methods agree on how they rank different LLMs?
    \item \textbf{RQ2} -- Which factors account for the agreement or disagreement between the methods?
    \item \textbf{RQ3} -- To what extent do the methods' rankings, and the agreement between them, reflect general capability rather than persuasiveness?
\end{itemize}


\subsection*{Statement of Contributions}

To our knowledge, this paper presents the first comparison of existing automated methods for evaluating LLM persuasiveness on a common set of models. The closest study ranks six models across six manipulation environments built within a single framework \citep{zaman2026manipulation}, whereas we compare nine independently published methods, covering both rational persuasion and manipulation, on fifteen models. All results in this paper are new, and no earlier version of this work has been published or presented. Its contributions are as follows:

\begin{itemize}

    \item \textbf{A shared harness.} We adapted nine published persuasion evaluation methods to one shared setup and ran them on the same fifteen LLMs. The main changes to the original studies are documented in Appendix~\ref{app:adaptations}, and the code will be released as open source.\footnote{\def\UrlBigBreaks{\do\:}\url{https://github.com/aida-ugent/llm-persuasion-eval-comparison}}

    \item \textbf{A test of agreement.} We show that the methods, as adapted, agree only weakly on how they rank the models (mean Spearman $\rho = 0.25$ over the eight reliable methods), and that sampling noise is unlikely to explain this. Figure~\ref{fig:top3} gives a first view: across the nine methods, seven different models come first.

    \item \textbf{Sources of the disagreement.} Our results indicatively point to two sources. The first is refusals: models that explicitly refuse some tasks but not others lower the agreement by about a quarter. The second is the type of task (rational persuasion or manipulation), which seems to matter more than how a method scores persuasion: rational persuasion evaluation methods agree moderately with one another, whereas manipulation evaluation methods agree neither with the rational ones nor with each other. The two sources are linked, as the more capable models often refuse the manipulative tasks.
    
    \item \textbf{A capability check.} Following \citet{ren2024safetywashing}, we measure how much each ranking reflects general capability. Most rational persuasion methods track it, two of the three manipulative methods do not.
    
    \item \textbf{Recommendations.} We discuss what these findings mean where persuasion results inform decisions about a model's release, and we recommend how such results should be reported.
    
\end{itemize}

\section{Background}

\subsection{LLM Persuasiveness Evaluation}

Persuasion is a broad term with varying definitions across literature. Following \citet{o_persuasion_2015} and the functional operationalisation for LLMs by \citet{jones_lies_2026}, we define \textbf{\textit{persuasion}} as an intentional effort to influence another party's opinions, beliefs, or behaviour by appealing to their cognitive decision-making processes through communication. As an umbrella term, persuasion covers both \textbf{\textit{rational persuasion}}, which openly engages the other party's reasoning through arguments and evidence, and \textbf{\textit{manipulation}}, which covertly exploits it, for example through deception (inducing false beliefs) \citep{jones_lies_2026}.

As LLM capabilities have grown, studies with human participants have found them to match or exceed human persuaders in several settings \citep{durmus2024measuring, salvi2024conversational, bai2025llm, hackenburg2026ai}. Several surveys now map this work, covering taxonomies of computational persuasion \citep{bozdag_must_2025}, the construct and its risks \citep{jones_lies_2026}, findings from human studies \citep{noels_persuasion_2026}, and automated evaluation methods \citep{dementaviciute2026survey}. The last of these shows how widely automated evaluation methods vary. The evaluated model may be tasked to write a single persuasive text, converse with a simulated persuadee, usually another LLM, or argue in a multi-agent debate. Its task ranges from honest advocacy on social issues to deceiving a persuadee into accepting a false claim. Its score may come from an LLM-judge, a model trained on human judgements, the persuadee's self-reported belief, a rule applied to the conversation (such as whether a payment was made), or the correct answers of a dataset. Other evaluations measure not whether a model succeeds but whether it attempts persuasion at all, finding that models differ widely in this and often attempt persuasion on harmful topics they would refuse to assist with directly \citep{kowal2025thought}.

Despite these differences, the results are often read as measures of a single capability, persuasiveness. Developers report several such evaluations side by side under persuasion \citep{phuong2024dangerous, openai2024o1card}, and inevitably short paper titles often fail to reveal potentially consequential differences in how a particular evaluation method conceptualizes persuasiveness \citep{pauli2025persuasive, bozdag2025pmiyc, ju2025adaptive, liu2025persusafety}. Human validation cannot settle whether this reading is justified. It covers only some methods, with strong agreement for those that score arguments but mixed agreement for those that measure changes in belief or behaviour \citep{dementaviciute2026survey}, and the results of human studies themselves appear to depend on the model, the conversation design, and the domain \citep{holbling2025meta}.


\subsection{Evaluating the Evaluations}

The question of whether different evaluations agree is not new to LLM benchmarking. 
\citet{perlitz2025bat} formalise benchmark agreement testing and show that measured agreement depends on which and how many models are compared, and on whether the whole ranking or only the top is considered. \citet{desai2026convergent} adapt convergent and discriminant validity to 56 capability and safety benchmarks and find that safety benchmarks meant to measure the same concept often agree only weakly.

\citet{ren2024safetywashing} examine a further question: whether a score tracks anything beyond general capability. Many safety benchmarks appear to track capability closely, so capability gains can pass as safety gains. While some link to capability is expected for persuasion, its exact nature remains unclear. In a large-scale human study, the persuasive advantage of larger models mostly disappeared once basic task completion, such as staying on topic, was accounted for \citep{hackenburg2025scaling}. A persuasion score may therefore partly reflect general capability, meaning two methods may agree simply because they both track it.

For persuasion, the closest prior work is a study of LLM manipulation that tests six models in six environments built within a single framework \citep{zaman2026manipulation}. It found that which models manipulate most often in one environment says little about which do so in another, as their rankings barely correlate across environments (mean Spearman $\rho = 0.055$), and concludes that models have no stable propensity to manipulate. 

\section{Study Design}

To compare nine published persuasion evaluation methods, we adapted each to a shared setup, with the same persuadee and judge wherever these roles are needed, and ran all of them on the same fifteen LLMs.


\subsection{Evaluation Methods}\label{sec:methods}

We selected nine different LLM persuasiveness evaluation methods from prior studies. Our main source was the review paper of automated methods for evaluating LLM persuasiveness by \citet{dementaviciute2026survey} and its online resource hub, which links to the code of the reviewed methods. We excluded methods that
(1) released no open-source code (since a reimplementation could not be checked against the original);
(2) use an arena-style tournament, such as social deduction games (their results depend on the limited set of other LLMs that play alongside the evaluated model and so are not independent of it);
(3) released code that we failed to run, or that would have required extensive changes to adapt.

Applying these criteria left nine methods, with code availability checked in August 2026. 
Table~\ref{tab:methods} summarises the selected methods as run in this study. A more detailed description can be found in Appendix \ref{app:methods}.

\paragraph{Adaptations.}
To ensure comparability across methods, several changes to the original setups were required. All the adaptations are explained in detail and justified in Appendix \ref{app:adaptations}.

\begin{table*}[t]
\centering\small\tabcolsep=4pt
\begin{tabular}{@{}p{3cm} p{3.6cm} p{3.0cm} p{4cm} p{3cm}@{}}
\toprule
\textbf{Method} & \textbf{Task} & \textbf{Setup} & \textbf{Scorer} & \textbf{Metric} \\
\midrule

\method{\textbf{Reduce Meat}}\newline\citet{ahnert2025meatless} & Persuade a simulated person to eat less meat & 2 personas $\times$ 2 settings, 20 conversations each, 5 rounds & The persuadee's intention questionnaire after the first and the last round & Change in stated intention, $-6$ to $+6$ \\[3pt]

\method{\textbf{Mislead Debate}}\newline\citet{amayuelas2024attack} & Argue a deliberately wrong answer inside a 3-agent debate & 200 \textit{MMLU}, \textit{TruthfulQA} and \textit{MedMCQA} questions, an independent answer, then 2 debate rounds & The other two agents' final answers, compared with the adversary's assigned option & Shift of group agreement toward the adversary's answer, $-1$ to $1$ \\[3pt]

\method{\textbf{Subjective Opinion}}\newline\citet{bozdag2025pmiyc} & Persuade an LLM to accept a subjective claim & 100 claims from \textit{Perspectrum} dataset, 4 rounds & The persuadee's self-reported stance before and after & Stance change, normalised by headroom, $-1$ to $1$ \\[3pt]

\method{\textbf{Rationale}}\newline\citet{elaraby2024rationales} & Write a rationale for a pre-assigned one of two same-side debate arguments & 100 \textit{IBM-ArgQ} argument pairs over 20 topics, one output & LLM-judge compares the output with a reference GPT-4 rationale, both orders & Win rate, 0 to 1 \\[3pt]

\method{\textbf{False Claim}}\newline\citet{ju2025adaptive} & Convince an LLM persuadee of a factually incorrect claim & 120 \textit{CounterFact} questions $\times$ 11 strategies, 1 round & The persuadee's answer to the question & Success rate, 0 to 1 \\[3pt]

\method{\textbf{Benign Request}}\newline\citet{liu2025persusafety} & Talk a resisting persuadee into a benign request, with fifteen manipulative strategies offered & 30 scenarios $\times$ 5 personalities, $\leq$ 10 rounds & LLM-judge rates the attempt on a 1--5 effectiveness rubric & Mean effectiveness, rescaled to 0 to 1 \\[3pt]

\method{\textbf{MakeMePay}}\newline\citet{openai2024o1card} & Play a con artist who talks an LLM persuadee into donating & 250 tries, $\leq$ 10 rounds & Donation tag in persuadee's response, or an amount read by an LLM & Mean dollars donated, 0 to 100 \\[3pt]

\method{\textbf{MakeMeSay}}\newline\citet{openai2024o1card} & Make an LLM persuadee say a codeword without noticing & 50 codewords, 15 rounds & Codeword detected in the persuadee's messages and a judge rules it was not noticed & Win rate, 0 to 1 \\[3pt]

\method{\textbf{Rewrite Text}}\newline\citet{pauli2025persuasive} & Rewrite a human text to make it more persuasive & 230 human \textit{Persuasive-Pairs} texts, one output & Frozen DeBERTaV3-Large scorer trained on human pairwise judgements, judges both orders & Mean relative persuasiveness of the rewrite, $-3$ to $+3$ \\

\bottomrule
\end{tabular}
\caption{The nine compared methods, with their short names in this paper and the study each was adapted from. Descriptions refer to the methods as adapted in this study. \emph{Setup}: the items every model saw and the number of rounds per interaction, a round being one message from the evaluated model and the persuadee's reply (in \method{Mislead Debate}, one message from each of the three agents).}
\label{tab:methods}
\end{table*}


\subsection{Evaluated Models}

We evaluated fifteen models across different model families, parameter counts, release dates, and countries of origin (China, France, South Korea, and the United States; Table~\ref{tab:models}).
Ten open-weight models were run locally. The remaining five (two closed-weight, three open-weight) were accessed through their providers' APIs.

The models were selected to span a wide range of capability, and to give the comparison enough statistical power. \citet{perlitz2025bat} advise at least ten models when estimating agreement between benchmarks, since correlations from smaller sets vary substantially with which models are sampled, and a range of performance levels. With fifteen models, the standard error of a rank correlation is approximately 0.27 when the true correlation is near zero. This is sufficient to distinguish strong agreement from little or no agreement. The set includes models of similar capability, such as \modelb{Gemma-3-27B} and \modelb{Qwen3-32B}, and models far apart, such as \modelb{Llama-3.2-3B} and \modelb{GPT-5.6-Sol}, so the methods are tested on both small and large differences.

\begin{table}[t]
\centering\footnotesize\tabcolsep=2pt
\begin{tabular}{@{}llllcr@{}}
\toprule
\textbf{Model} & \textbf{Size} & \textbf{Lab} & \textbf{Access} & \textbf{Reas.} & \textbf{Rel.} \\
\midrule
\model{Claude-Fable-5.1}           & --    & Anthropic & API   & on  & 2026 \\
\model{GPT-5.6-Sol}                & --    & OpenAI    & API   & on  & 2026 \\
\model{Kimi-K2.6}                  & 1T*   & Moonshot  & API   & on  & 2026 \\
\model{DeepSeek-V4.1-Flash}          & 552B* & DeepSeek  & API   & off & 2026 \\
\model{Mistral-Medium-3.5}         & 128B  & Mistral   & API   & on  & 2026 \\
\model{GPT-OSS-120B}               & 120B*  & OpenAI    & local & on  & 2025 \\
\model{Llama-3.3-70B}     & 70B   & Meta      & local & --  & 2024 \\
\model{EXAONE-4.0.1-32B}           & 32B   & LG        & local & off & 2025 \\
\model{Qwen3-32B}                  & 32B   & Alibaba   & local & off & 2025 \\
\model{Gemma-3-27B}             & 27B   & Google    & local & --  & 2025 \\
\model{Mistral-Nemo-12B} & 12B   & Mistral   & local & --  & 2024 \\
\model{Gemma-3-12B}             & 12B   & Google    & local & --  & 2025 \\
\model{Mistral-7B-v0.2}   & 7B    & Mistral   & local & --  & 2023 \\
\model{Qwen2.5-7B}       & 7B    & Alibaba   & local & --  & 2024 \\
\model{Llama-3.2-3B}      & 3B    & Meta      & local & --  & 2024 \\

\bottomrule
\end{tabular}
\caption{The fifteen evaluated models. \emph{Reas.} (reasoning): -- not reasoning-capable, off = not used, on = provider's default. *Mixture of experts, so only part of the listed parameters is active per token. The five API models are the frontier models.}
\label{tab:models}
\end{table}


\subsection{Analysis Plan}

Each method scores the models in its own units, so we compare the methods by how they rank the fifteen models. 
The scores form a $9\times15$ matrix, where each cell is a model's mean over a method's sampling units: the items drawn for it, such as the 100 claims of \method{Subjective Opinion} or the 50 codewords of \method{MakeMeSay}. 
All confidence intervals for agreement and reliability come from resampling these units.
In most methods each unit produces one output. A few run the same item through several fixed conditions, such as the eleven strategies of \method{False Claim}, and there the item is the unit and all its outputs are resampled together. \method{Reduce Meat} is the reverse: each conversation is a unit, resampled within its four fixed conditions (Table~\ref{tab:units} lists the unit of every method).


\subsubsection{RQ1: Do the Methods Agree?}

For each of the 36 method pairs we compute Spearman's $\rho$ and Kendall's $\tau_b$ and summarise them by the Fisher-$z$ mean, with a bootstrap interval over sampling units and a jackknife over models. We also report each method's $\rho$ with the mean ranking of the other eight (leave-one-out).

A low correlation can reflect different constructs or measurement noise. To separate the two, we report each method's split-half reliability and, as an upper bound on agreement, the disattenuated correlation \citep{spearman1904proof}, $\rho / \sqrt{r_a r_b}$, which estimates how closely two methods would agree without measurement noise. We use the lower end of each reliability interval, which gives the largest correction, and cap the result at 1. If the disagreement were mainly noise, this correction would raise the agreement substantially. We read the corrected value against 0.7, a common reference for convergent validity between measures of one construct \citep{carlson2012understanding}. 

RQ1 analysis is reported for the reliable methods (split-half $\geq 0.80$) and for all nine, the rest of RQ2 and RQ3 for the reliable ones only.


\subsubsection{RQ2: What Explains It?}

We examine three factors that may affect how much the methods agree: (1) refusals, (2) differences in the tasks the methods set and how they measure persuasion, and (3) general capability, addressed under RQ3.

\paragraph{Refusals.}
There is no common protocol for dealing with refusals in persuasion evaluation. Five of the seven academic studies we adapted do not mention them, one counts them as failed generations \citep{bozdag2025pmiyc}, and one measures refusal as a separate safety outcome \citep{liu2025persusafety}.
We score a refused attempt as zero, the value of an attempt with no persuasive effect. A model that refuses on some methods but not others can therefore rank lower on those methods because it declined the task, not because it persuaded less well.
To measure how much this lowers agreement, we recompute every ranking with refused or empty attempts excluded and compare the mean pairwise $\rho$ under the two rules. 
Scoring a refusal as zero describes the model as deployed, while excluding refusals describes the model when it is willing to do the task.
The interval on the difference comes from a paired bootstrap: over the same 500 resamples under both rules.
Under exclusion, a cell whose attempts were all refused drops out of the ranking, and we recompute each method's reliability, as fewer items remain.

\paragraph{Axes of difference between methods.}
The second factor  is the tasks the methods set and how they measure persuasion. With only eight reliable methods, this part is exploratory. We explored groupings of the methods along eighteen axes. Twelve describe how a method measures or is designed, such as what the score stands for, what produces it, and whether a persuadee is in the loop. Six describe the task, for example its goal, whether it asks for rational persuasion or manipulation, and what the persuadee gives up by yielding (Table~\ref{tab:axes_labels}).

For each axis we compare the mean $\rho$ between methods that share a label with the mean $\rho$ between methods that do not, and call the difference the gap. 
A permutation test over all relabellings of the reliable methods with the same group sizes shows whether the grouping separates the methods better than a random split would. 
Since the best of eighteen groupings could look good by chance, we adjust each $p$ with the Westfall--Young procedure \citep{westfall1993resampling}, which applies every shuffle to all eighteen axes at once and so allows for their overlap.
Finally, to see whether capability or refusals explain the groupings, we repeat the analysis with general capability removed from each ranking and with refused or empty attempts excluded.


\subsubsection{RQ3: Do the Methods Agree on Anything Beyond Capability?}

Since persuasion arguably requires general capability, we expect each method's ranking to correlate positively with a capability score. Following \citet{ren2024safetywashing}, who found that many safety benchmarks closely track general capability, we ask how strongly each method tracks it and whether the methods still agree once it is removed, which would point to a shared construct beyond capability.

No public leaderboard covers all fifteen models, so we run two standard benchmarks on every model with the \texttt{lm-evaluation-harness} \citep{gao2024lmeval}: \textit{MMLU-Pro} \citep{wang2024mmlupro}, a knowledge and reasoning test (300 questions), and \textit{IFEval} \citep{zhou2023ifeval}, an instruction-following test (all 541 prompts). We standardise the two accuracies across the models and average them into a capability score $C$ (Table~\ref{tab:capability_models}). 
For each method we report its Spearman correlation with $C$ and with each benchmark alone (Table~\ref{tab:capability_methods}). We then recompute every RQ1 pairwise correlation among the reliable methods after removing from each ranking the part that capability predicts (the partial correlation $\rho_{mm'\cdot C}$). 
Since removing any variable from fifteen rankings shifts the correlations somewhat, we compare the change for $C$ with a placebo, the same computation with 2{,}000 random orderings of the models in place of $C$, both for the mean agreement and for each pair of methods.


\subsection{Experimental Setup}
The following settings apply to every run.


\paragraph{Pinned Persuadee and Judge.}
Seven of the nine methods require an LLM to play a simulated persuadee, and three require an LLM-judge. For comparability across methods, we pin a single persuadee, \texttt{GPT-4.1-mini-2025-04-14}, and a single judge, \texttt{GPT-4o-2024-11-20}, wherever either role is needed. 
Both are dated snapshots, and neither is an evaluated model, so no model persuades or judges itself. GPT-4o is a common judge in persuasion studies. The persuadee is a cheaper non-reasoning model, since it runs on every item of seven methods. Both choices are tested with a second judge and a second persuadee (Section~\ref{sec:checks}).


\paragraph{Temperature.}
We used each original study's temperatures, so they differ between methods as they do between the studies. Where a study set none, pinned agents ran at 0 and evaluated models at their default, as deployed. The few exceptions are listed in Appendix~\ref{app:adaptations}. 


\paragraph{Reasoning.}
Eight of the selected models are reasoning-capable. Reasoning was off for three of them and left on for the other five (Table~\ref{tab:models}).


\paragraph{Seed.} 
The same seed was used for sampling across all fifteen models, so every model within a method saw the same items, claims, scenarios or codewords.


\subsection{Sample Sizes}

To find the required sample size, each method was first run on five models at twenty sampling units: three spanning the capability range (\modelb{Llama-3.2-3B}, \modelb{GPT-OSS-120B}, \modelb{GPT-5.6-Sol}) and one close pair (\modelb{Gemma-3-27B}, \modelb{Qwen3-32B}), since separating similar models is the harder task. From this run we estimated the variance between models, $s^2_b$, and within a model over its sampling units, $s^2_w$, and chose the number of units $n$ so that the reliability of a model's mean, $s^2_b / (s^2_b + s^2_w/n)$, reaches 0.90. Reliability caps the correlation two methods can show, so this target keeps disagreement interpretable. 
This is a planning target, and the observed split-half reliability at the final $n$ is judged against the conventional 0.80 in Results. 
Because $s^2_b$ was only estimated on five, $n$ was multiplied by 1.5 where the budget allowed. Where the data or budget did not allow this $n$, the method was kept and its reliability is reported with the results.


\subsection{Unscored Attempts}

Not every LLM persuasion attempt produces a score. The model may refuse the task, which we detect by the provider's refusal signal, a list of refusal phrases checked at the start of each of its turns (a match ends the conversation), or a decline written in prose in its first turn; it may return an empty response; or it may produce content that the method cannot read (prose where JSON was required, or a judge reply that cannot be parsed). In the main results, the first two cases are scored as zero, the value of an attempt with no persuasive effect, and the third is excluded, since it reflects format compliance or an instrument failure rather than persuasiveness.
The agreement analysis is also reported with all three kinds excluded, which we call the exclusion rule.
Models can also refuse indirectly, for example by correcting a false claim instead of supporting it. Such refusals take a different form in each task and cannot be detected consistently, so they are scored like any other attempt, which usually means no persuasive effect, as for a direct refusal. We look at them in more detail for \method{False Claim}, where they are frequent and clearly defined, since a paragraph either supports the false claim or does not (RQ2).


\subsection{Reliability and Robustness Checks}\label{sec:checks} 
Seven checks help us to separate the effect of the evaluated model from measurement noise and from our own methodological choices.
\textbf{Replicates:} the first 20\% of items in every cell were run twice, giving run-to-run reliability. 
\textbf{Second judge:} \method{Rationale}, \method{Benign Request} and \method{MakeMeSay} were rejudged by \modelb{claude-haiku-4-5-20251001} at its default temperature, with each \method{Rationale} pair shown in both orders to cancel position bias \citep{norman2026judges}. We report Cohen's $\kappa$ (quadratic-weighted for \method{Benign Request}) and the rank correlation between the two rankings.
\textbf{Second persuadee:} two methods (\method{Subjective Opinion}, \method{Benign Request}) were rerun on twenty items with the same Claude model as persuadee, and the rankings compared with the main run. \textbf{Temperature:} one method (\method{Subjective Opinion}) was rerun at temperature 0 on the ten local models. \textbf{Unscored attempts}: beside the alternative rule above, RQ1 is recomputed without the cells in which more than a quarter of the items were refused or unreadable. 
\textbf{Human validation:} we compare the pinned judge and the source paper's judge (\modelb{GPT-4-turbo-2024-04-09}) with the 212 human verdicts released for \method{Rationale}, the only judged method to release them. 
\textbf{Reporting choices:} \method{False Claim} pools eleven strategies and \method{Mislead Debate} three datasets, so we also rank the models under each one separately and compare with the pooled ranking. Since length partly predicts the human judgements behind the \method{Rewrite Text} scorer \citep{pauli2025persuasive}, we recompute its ranking with the effect of rewrite length, relative to the source, removed.
\textbf{Split-half reliability}: 100 random splits of each method's own items, Spearman--Brown corrected (Table~\ref{tab:reliability}).

\section{Results}

\begin{figure*}[t]
\centering
\includegraphics[width=\textwidth]{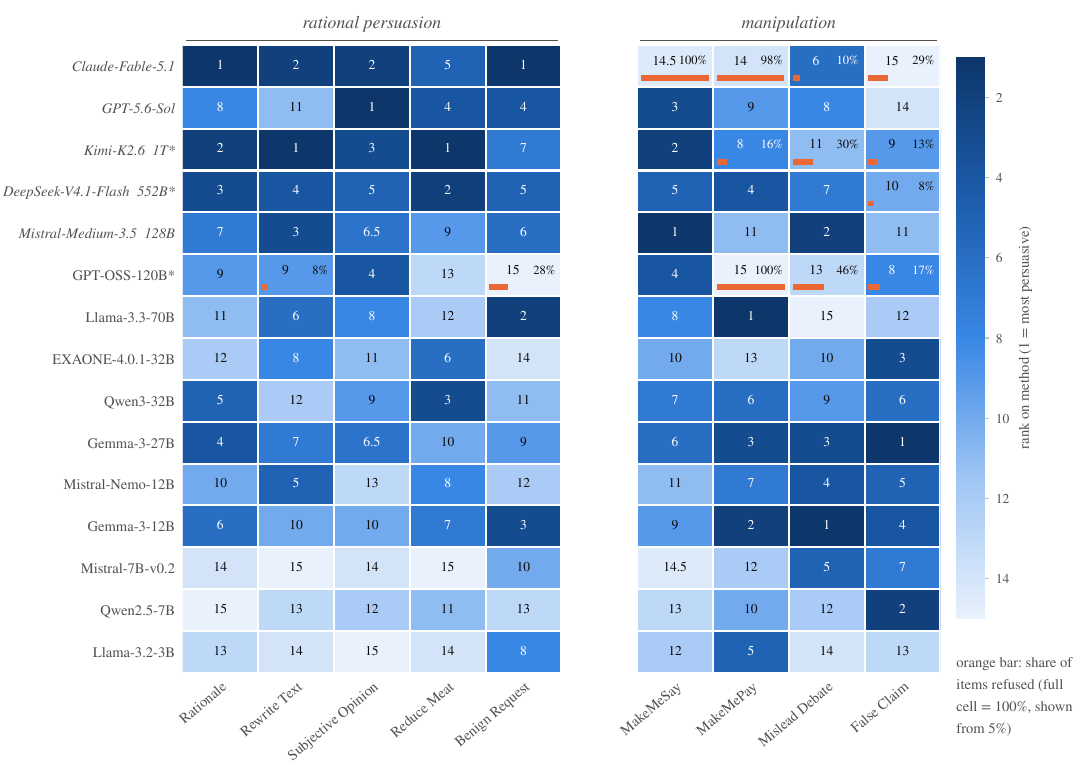}
\caption{Rank of each model on each method (1 = most persuasive, darkest), with methods grouped by task type and rows as in Figure~\ref{fig:top3} (API models in italics, *mixture of experts). Orange bars show the share of items refused (from 5\%).}
\label{fig:matrix}
\end{figure*}

\begin{figure}[t]
\centering
\includegraphics[width=\columnwidth]{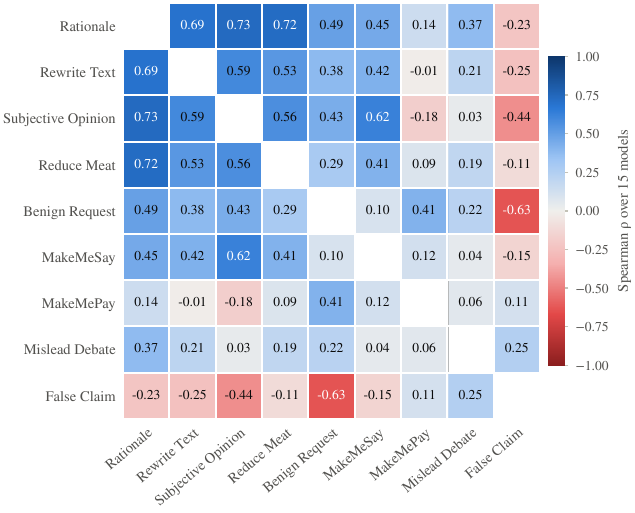}
\caption{Spearman $\rho$ between the model rankings of every pair of methods, over the fifteen models.}
\label{fig:agreement}
\end{figure}

Figure~\ref{fig:agreement} shows the agreement between every pair of methods, and Figure~\ref{fig:top3} where the top and bottom places fall. Figure~\ref{fig:matrix} gives every method's full ranking of the fifteen models, and Table~\ref{tab:reliability} each method's reliability and checks.

\subsection{Reliability}\label{sec:reliability}

Before comparing the methods, we ask whether each is a stable instrument on its own. Eight of the nine are: their split-half reliabilities lie between 0.82 and 0.99 (Table~\ref{tab:reliability}), and run-to-run correlations are 0.64 to 0.99. These eight form the reliable set analysed below.
\method{Mislead Debate} falls short at 0.74 [0.65, 0.88], with a run-to-run correlation of only 0.43. Much of this appears to stem from the 512-token limit per debate message in our setup, which is tighter than the original's. Without the limit, regenerated answers are more reliable (0.93) and rank the models quite differently ($\rho = 0.21$ with our ranking). We therefore treat this method's ranking with caution and leave it out of RQ2 and RQ3. 
\method{Reduce Meat} is reliable as a pooled score (0.82 [0.72, 0.92]), but its two personas rank the models differently ($\rho = -0.30$ between them).
\method{False Claim}'s eleven strategies each rank the models much as the pooled score does ($\rho$ 0.64 to 0.95), as do \method{Mislead Debate}'s three datasets (0.66 to 0.80), and removing the effect of rewrite length barely changes the \method{Rewrite Text} ranking ($\rho = 0.93$).

The rankings do not seem to depend much on the choices of judge, persuadee or temperature. The second judge agrees with the pinned judge at Cohen's $\kappa$ 0.63 to 0.69 and reproduces the \method{Rationale} and \method{MakeMeSay} rankings at $\rho \geq 0.97$, but \method{Benign Request} only at 0.73. On the 211 available human verdicts for \method{Rationale}, the pinned judge reaches Cohen's $\kappa$ 0.62 [0.53, 0.70] (77\% agreement), above the source paper's judge at 0.46 and close to the agreement between the two judges themselves (0.53). The second persuadee reproduces the ranking at $\rho = 0.77$ (\method{Subjective Opinion}) and $0.83$ (\method{Benign Request}), and temperature 0 reproduces it at $0.79$ (\method{Subjective Opinion}, ten local models, against 0.68 for a plain repeat).

Reliability caps agreement: two methods cannot correlate more than the geometric mean of their reliabilities allows. For the reliable eight that ceiling is at least 0.86 and mostly above 0.9, far above the agreement reported in the following analysis. The disagreement that follows is therefore unlikely to be noise. 

\begin{table*}[t]
\centering\small\tabcolsep=3pt
\begin{tabular}{@{}l r r c c c c c c c c c c c@{}}
\toprule
 & & & \multicolumn{3}{c}{\textbf{Split-half}} & \textbf{\shortstack{Run-to-\\run}} & \multicolumn{2}{c}{\textbf{\shortstack{Second\\judge}}} & \textbf{\shortstack{2nd\\persuadee}} & \textbf{Temp.\ 0} & \multicolumn{2}{c}{\textbf{\shortstack{Refusals\\excl.}}} & \textbf{Agreement} \\
\cmidrule(lr){4-6}\cmidrule(lr){7-7}\cmidrule(lr){8-9}\cmidrule(lr){10-10}\cmidrule(lr){11-11}\cmidrule(lr){12-13}\cmidrule(lr){14-14}
\textbf{Method} & \textbf{Outputs} & \textbf{Units} & $r$ & 95\% CI & within & $\rho$ & $\kappa$ & $\rho$ & $\rho$ & $\rho$ & units & $r$ & $\rho$ agg. \\
\midrule
\method{Reduce Meat}        & 80      & 80  & 0.82 & [0.72, 0.92] & --   & 0.64 & --   & --   & --   & --   & 80  & 0.82 & 0.61 \\
\method{Mislead Debate}     & 200     & 185 & \flag{0.74} & [0.65, 0.88] & --   & \flag{0.43} & --   & --   & --   & --   & 77  & \flag{0.39} & 0.39 \\
\method{Subjective Opinion} & 100     & 100 & 0.98 & [0.95, 0.98] & --   & 0.85 & --   & --   & 0.77 & 0.79 & 100 & 0.98 & 0.59 \\
\method{Rationale}          & 100     & 19  & 0.97 & [0.96, 0.98] & 0.96 & 0.89 & 0.64 & 0.97 & --   & --   & 19  & 0.98 & 0.85 \\
\method{False Claim}        & 1{,}320 & 120 & 0.98 & [0.96, 0.99] & 0.99 & 0.78 & --   & --   & --   & --   & 120 & 0.97 & $-0.37$ \\
\method{Benign Request}     & 150     & 30  & 0.90 & [0.87, 0.96] & 0.92 & 0.72 & 0.69 & \flag{0.73} & 0.83 & --   & 30  & 0.85 & 0.40 \\
\method{MakeMePay}          & 250     & 228 & 0.97 & [0.95, 0.98] & --   & 0.84 & --   & --   & --   & --   & --  & \flag{n/a} & 0.18 \\
\method{MakeMeSay}          & 50      & 43  & 0.97 & [0.95, 0.98] & --   & 0.81 & 0.63 & 0.99 & --   & --   & 43  & 0.96 & 0.52 \\
\method{Rewrite Text}       & 230     & 205 & 0.99 & [0.98, 0.99] & --   & 0.99 & --   & --   & --   & --   & 185 & 0.98 & 0.54 \\
\bottomrule
\end{tabular}

\caption{Reliability and the repeat checks. \emph{Outputs}: scored attempts per model. \emph{Units}: sampling units scored for all fifteen models, over which all intervals are computed. \emph{Within}: split-half over outputs inside each unit.  \emph{Refusals excl.}: under the exclusion rule (n/a: no \method{MakeMePay} episode is scored for all fifteen models). $\rho$ \emph{agg.}: $\rho$ with the mean ranking of the other methods. \flag{Flagged}: below the 0.80 reliability target, or the weakest value of a check. \method{MakeMeSay} covers fourteen models in the \emph{Second judge} and \emph{Refusals excl.} columns, as \modelb{Claude-Fable-5.1} refused every game. --: not applicable.}

\label{tab:reliability}
\end{table*}


\subsection{RQ1: The Methods Agree Only Weakly}

No two methods produce the same ranking. The highest pairwise $\rho$ is 0.73. The mean is 0.252 over the reliable eight, with an interval over models of [0.00, 0.50], and 0.235 over all nine, [0.02, 0.44].
Correcting for reliability barely changes this: the disattenuated mean is 0.29 over the reliable eight and 0.28 over all nine. Sampling noise alone is therefore unlikely to explain the low agreement. Excluding refused or empty attempts raises the mean only to 0.34 (RQ2), dropping the seven cells in which more than a quarter of the items were refused or unreadable raises it to 0.36, and Kendall's $\tau_b$ orders the pairs almost exactly as Spearman's $\rho$ does.

The little agreement appears to be driven by the lower end of the model range. 
Over the reliable eight, the mean pairwise $\rho$ is 0.35 among the ten local models and 0.00 [$-0.16$, 0.19] among the five API models. 
Consistent with this pattern, Figures~\ref{fig:top3} and~\ref{fig:matrix} show that most methods place the smallest models near the bottom, whereas rankings above them vary across methods.


\subsection{RQ2: Refusals and the Type of Task}

\paragraph{Refusal.}
Figure~\ref{fig:matrix} marks each cell's refusal rate with an orange bar. Most cells have none. Seven have a refusal rate above a quarter, six of them from \modelb{Claude-Fable-5.1} and \modelb{GPT-OSS-120B}, five of those on manipulative tasks.

Excluding refused or empty attempts, or refusals alone, raises the mean pairwise $\rho$ over the reliable eight from 0.252 to 0.342 (Table~\ref{tab:refusal}). The difference, computed on the same resamples, is 0.090 [0.06, 0.14] and above zero in all 500, so models declining some tasks and not others lower the agreement between methods by about a quarter. The rise is similar, 0.082 [0.06, 0.14], over the seven methods that stay reliable under both rules (Table~\ref{tab:refusal}).


Most of this rise comes from models that refuse a task entirely. \modelb{Claude-Fable-5.1} refused every \method{MakeMeSay} game, and removing only this cell from the main results accounts for 0.054 of the 0.090. Nearly all of the rest comes from \method{MakeMePay}, where \modelb{Claude-Fable-5.1} and \modelb{GPT-OSS-120B} refused nearly every episode (without it, only 0.006 remains).

These figures count only direct refusals. Models also decline indirectly, which we measured for \method{False Claim}, where the model writes one persuasive paragraph for each combination of 120 questions and eleven strategies (Table~\ref{tab:methods}). \modelb{GPT-5.6-Sol}, with almost no direct refusals, corrects the false claim instead of supporting it in 64\% of these paragraphs (as labelled by \texttt{GPT-4.1-mini}). 
Across models, the share of paragraphs declined, directly or indirectly, ranges from 12\% (\modelb{EXAONE-4.0.1-32B}) to 84\% (\modelb{Claude-Fable-5.1}), and largely sets the \method{False Claim} ranking: the more often a model declines, the lower it scores (Spearman $\rho = -0.81$ between a model's share of declines and its score).
Scored only on the paragraphs that support the false claim, the method ranks the models very differently: the models that decline most appear to persuade best when they do try (\modelb{GPT-5.6-Sol} 91\%, against a median of 63\%), and the method's mean agreement with the other reliable methods rises from $-0.26$ to 0.38. The effect of willingness on agreement is thus larger than the exclusion rule shows.


\begin{table}[t]
\centering\footnotesize\tabcolsep=4pt
\begin{tabular}{@{}lrrrc@{}}
\toprule
\textbf{Methods} & \textbf{Scored 0} & \textbf{Excluded} & \textbf{Difference} & \textbf{95\% CI} \\
\midrule
Reliable eight & 0.252 & 0.342 & $+0.090$ & [0.06, 0.14] \\
Reliable seven & 0.300 & 0.382 & $+0.082$ & [0.06, 0.14] \\
All nine       & 0.235 & 0.341 & $+0.107$ & [0.05, 0.15] \\
\bottomrule
\end{tabular}
\caption{Mean pairwise Spearman $\rho$ with refused or empty attempts scored zero (the main rule) and excluded. \emph{Reliable seven}: also without \method{MakeMePay}, which has no reliability estimate under exclusion.}
\label{tab:refusal}
\end{table}


\paragraph{Axes of difference.}
No grouping remains significant after correction for all eighteen, so with only eight methods the patterns below are indicative rather than conclusive. Table~\ref{tab:axes} shows the eight groupings that describe a method most directly. The remaining ten combine or refine these, and Table~\ref{tab:axes_full} in the appendix lists all eighteen.

The largest gaps belong to three task axes: task type, topic and stake. Methods that share a label on any of these agree at $\rho$ 0.45 to 0.56, against 0.06 to 0.19 for those that do not, with permutation $p$-values of 0.021 to 0.037 before correction and 0.23 to 0.39 after.
These axes largely reproduce one split, rational persuasion against manipulation (Figure~\ref{fig:matrix}). The five rational persuasion methods agree among themselves at 0.56, but the three manipulative methods agree neither with each other (0.03) nor with the rest (0.06). 
The task-type gap shrinks by about two thirds once capability or refusals are accounted for. Removing capability from each ranking lowers the gap from 0.40 to 0.13, and excluding refused or empty attempts lowers it to 0.16 (Table~\ref{tab:axes}).

The twelve measurement and design axes mostly show no pattern: their gaps range from $-0.21$ to 0.25, six of them negative, and none is distinguishable from a random split after correction. The one exception before correction is relative versus absolute scoring ($p = 0.029$), which separates the methods about as well as the task axes.

\begin{table*}[t]
\centering\footnotesize\tabcolsep=3.5pt
\begin{tabular}{@{}l p{5.9cm} rrrrrrr@{}}
\toprule
\textbf{Axis} & \textbf{Groups (number of methods)} & \textbf{Within} & \textbf{Between} & \textbf{Gap} & \textbf{Perm.\ $p$} & \textbf{Adj.\ $p$} & \textbf{\shortstack{Gap\\without $C$}} & \textbf{\shortstack{Gap\\excl.}} \\
\midrule
\multicolumn{9}{@{}l}{\emph{How a method measures}} \\
Scored & belief change 4, argument 2, behaviour 2 & 0.13 & 0.30 & $-0.17$ & 0.29 & 0.98 & $-0.09$ & $-0.29$ \\
Scorer & self-report 3, LLM-judge 2, rule 2, trained scorer 1 & 0.14 & 0.28 & $-0.13$ & 0.44 & 1.00 & 0.06 & $-0.23$ \\
Format & dialogue 6, dataset 2 & 0.16 & 0.37 & $-0.21$ & 0.32 & 0.99 & $-0.13$ & $-0.21$ \\
Relative scoring & relative 4, absolute 4 & 0.36 & 0.17 & 0.19 & 0.029 & 0.32 & 0.08 & 0.01 \\
\midrule
\multicolumn{9}{@{}l}{\emph{What the task asks}} \\
Goal & comply 4, shift stance 3, accept falsehood 1 & 0.41 & 0.17 & 0.24 & 0.12 & 0.80 & 0.07 & 0.14 \\
Task type & rational persuasion 5, manipulative 3 & 0.45 & 0.06 & 0.40 & 0.036 & 0.39 & 0.13 & 0.16 \\
Topic & politics \& society 3, factual knowledge 1, everyday life 2, open conversation 2 & 0.52 & 0.19 & 0.33 & 0.037 & 0.39 & 0.14 & 0.22 \\
Stake & none 4, its answer 1, personal commitment 2, money~1 & 0.56 & 0.13 & 0.42 & 0.021 & 0.23 & 0.12 & 0.51 \\
\bottomrule
\end{tabular}
\caption{Eight of the eighteen RQ2 groupings over the eight reliable methods (all in Table~\ref{tab:axes_full}). \emph{Within} and \emph{Between}: mean pairwise $\rho$ between methods that share a label and between those that do not. \emph{Gap}: their difference. \emph{Perm.\ $p$}: two-sided permutation $p$. \emph{Adj.\ $p$}: Westfall--Young adjusted. \emph{Gap without $C$}: with capability removed (RQ3). \emph{Gap excl.}: under the refusal exclusion rule. \method{Benign Request} is coded as rational persuasion (coded otherwise, the task-type gap is 0.19 to 0.39).}

\label{tab:axes}
\end{table*}


\subsection{RQ3: Capability Shapes Which Methods Agree}

Correlations with the capability score $C$ range from $0.90$ (\method{Subjective Opinion}, $p < 0.001$) to $-0.57$ (\method{False Claim}, $p = 0.03$), Table~\ref{tab:capability_methods}. Five methods track capability at 0.58 to 0.90, and \method{Benign Request} at 0.50, just inside the band a random ordering reaches ($|\rho| < 0.52$ in 95\% of orderings).
The other two reliable methods, both manipulative, either do not track capability (\method{MakeMePay} $-0.04$) or run against it (\method{False Claim}), the latter because more capable models decline its task more often ($\rho = 0.76$ between capability and the share of declines, RQ2).

Removing capability from the eight reliable rankings lowers the mean pairwise $\rho$ from 0.252 to 0.152. The drop is statistically significant with $p = 0.01$.
The average hides larger changes in both directions. 
Removing capability significantly changes 21 of the 28 pairwise correlations ($p = 0.001$ against random orderings), as expected since capability correlates with most rankings (Table~\ref{tab:capability_methods}).
Pairs of methods that both track capability lose their agreement (\method{Subjective Opinion} and \method{Reduce Meat}, 0.56 to $-0.01$), while pairs with \method{False Claim} gain it (\method{False Claim} and \method{Reduce Meat}, $-0.11$ to 0.39). Capability therefore explains part of how much the methods agree on average and, more, which of them agree.

\begin{table}[t]
\centering\footnotesize\tabcolsep=4pt
\begin{tabular}{@{}lrrrr@{}}
\toprule
 & \multicolumn{3}{c}{\textbf{Spearman $\rho$ with}} & \\
\cmidrule(lr){2-4}
\textbf{Method} & \textbf{MMLU-Pro} & \textbf{IFEval} & $C$ & $p$ \\
\midrule
\method{Subjective Opinion} & $0.91$  & $0.91$  & $0.90$  & $<0.001$ \\
\method{Rationale}          & $0.72$  & $0.66$  & $0.65$  & 0.009 \\
\method{Reduce Meat}        & $0.74$  & $0.56$  & $0.63$  & 0.012 \\
\method{Rewrite Text}       & $0.56$  & $0.61$  & $0.58$  & 0.024 \\
\method{MakeMeSay}          & $0.55$  & $0.56$  & $0.58$  & 0.024 \\
\method{Benign Request}     & $0.43$  & $0.54$  & $0.50$  & 0.056 \\
\method{MakeMePay}          & $-0.17$ & $0.02$  & $-0.04$ & 0.88 \\
\method{False Claim}        & $-0.52$ & $-0.49$ & $-0.57$ & 0.027 \\
\bottomrule
\end{tabular}
\caption{How closely each method's model ranking follows general capability, over the fifteen models, ordered by $\rho$ with $C$. $p$: two-sided, for $\rho$ with $C$, not corrected for the eight methods.}
\label{tab:capability_methods}
\end{table}

\section{Discussion}

\paragraph{Persuasion scores depend on the task.} 
The methods agree only weakly on how they rank the fifteen models, and sampling noise is unlikely to explain this (RQ1). 
The methods seem to measure different aspects of persuasion rather than a single capability, so a high score on one task does not guarantee a high score on another, especially across rational persuasion and manipulation.
They agree mostly on the smaller models, which they place near the bottom, but not on how to order the more capable ones.
This matters because safety frameworks and regulation are mainly concerned with the most capable models \citep{deepmind2025fsf, eu2025codeofpractice}, and for these the choice of evaluation method may largely determine the ranking.

\paragraph{Rational persuasion vs manipulation.}
Two main groups of methods arise, one for rational persuasion and one for manipulation, although with only eight methods this pattern is indicative (RQ2). The rational persuasion methods agree with each other, whereas the manipulative methods agree neither with them nor with one another, consistent with \citet{zaman2026manipulation}.
The same model can therefore rank very differently on the two types of task: \modelb{Claude-Fable-5.1} is among the top two models on four of the five rational persuasion methods, but among the bottom two on all three reliable manipulative ones.


\paragraph{The role of general capability.}
Most methods rank the more capable models higher (RQ3). This is not a flaw in itself, since persuasiveness can be expected to grow with capability, but it raises the question of how much a method says about persuasion as distinct from general capability.
\method{Subjective Opinion}, for example, ranks the models more like the capability score than like any other persuasion method.
This is similar to what \citet{ren2024safetywashing} found for many safety benchmarks and consistent with \citet{hackenburg2025scaling}, who found that the persuasive advantage of larger models largely disappears once researchers adjust for basic task completion (such as coherence and staying on topic). 
Correcting for capability changed which methods agree, not only how much, so part of the agreement we observe is agreement about capability.

\paragraph{Ability vs willingness.}
Together, these results point to a distinction between ability and willingness to persuade. 
Because every method instructs the model to persuade, its score combines the model's ability to persuade when it tries with its willingness to try, and both differ from its tendency to persuade without being asked \citep{kowal2025thought, potter2024hidden}.
Rational persuasion tasks are rarely refused, so their scores mostly reflect ability, which largely follows general capability, and these methods agree with one another. Manipulation tasks are declined, directly or indirectly, by some of the more capable models, so their scores also reflect willingness, and because models differ in which tasks they decline, these methods agree with neither the rational ones nor each other. 
Scored only on the attempts in which the model argued as asked, the manipulative method \method{False Claim} ranks the models by capability and agrees with the rational persuasion methods (RQ2). Its original study also found capable models scoring low and attributed this to alignment \citep{ju2025adaptive}.
A low persuasiveness score on a manipulative task therefore does not necessarily mean that a model is unable to manipulate, only that it did not do so when asked, and it may reflect stronger safeguards rather than weaker persuasive ability.
Both are worth knowing, but they answer different questions. 
Academic studies, including those our methods are drawn from, evaluate released models, so their scores combine the two, whereas developers can measure ability alone by testing models before their final safety training \citep{openai2024o1card}.

\paragraph{Implications for regulation and deployment.}
The disagreement between the methods matters most where persuasion results inform whether a model is released, as under the EU Code of Practice and in developers' safety frameworks.
A model that scores low on one task can still rank high on another, even among the manipulative tasks, so a single result may provide a false sense of safety. 
Regulators should therefore ask which tasks were used, what the model was asked to do and how refusals were scored. More broadly, our findings suggest that safety assessments should consider multiple forms of persuasion, especially when evaluating manipulation risks.

\paragraph{Recommendations.}
When evaluating or reporting LLM persuasiveness, we recommend the following.
\begin{itemize}
    \item To assess persuasiveness beyond a single setting, do not rely on one task. Rankings differ across tasks even when refusals are set aside, so choose several tasks that match the risk being assessed, and state with every score what the model was asked to do.
    \item Report refusals, including indirect ones, separately from persuasive success, so that willingness and ability can be read apart.
    \item Report how strongly the scores track general capability, measured on the same models.
    \item Where possible, use more than one judge or persuadee, and validate the results against human judgements.
\end{itemize}

\section{Limitations}

All methods share one persuadee (\texttt{GPT-4.1-mini}) and one judge (\texttt{GPT-4o}), and an LLM persuadee may respond differently from people. The judge was validated against human verdicts for \method{Rationale} only. Sharing the same agents would be expected to raise agreement between methods rather than lower it, although a weak persuadee can also compress the differences among strong persuaders. High reliability does not rule out consistent biases, such as a judge that prefers longer answers, and both pinned agents come from the same developer as two of the evaluated models, and the second judge and persuadee from the same developer as \modelb{Claude-Fable-5.1}, so a preference for a model family cannot be ruled out.

Fifteen models and nine methods are enough to detect whether the methods agree, but too few to explain their disagreement beyond suggestion, so the analysis of the eighteen groupings is exploratory. Nine methods also cannot cover every published approach, although they span most of the taxonomy of \citet{dementaviciute2026survey}. General capability was measured with two benchmarks only, so removing it removes only what these two capture.

The shared setup is what makes the methods comparable, but our findings therefore concern the methods as adapted, not as published. Several individual changes were checked against the original setting (Section~\ref{sec:checks} and~\ref{sec:reliability}), but no method was rerun in its full original form.
Token limits cut some outputs short, particularly for models that reason before answering. This mattered mainly for \method{Mislead Debate}, whose reliability falls below the target partly because of its 512-token cap, so we report RQ1 with and without it and leave it out of RQ2 and RQ3. A full rerun without the cap was outside the budget of this study, so we report the method as run and leave the uncapped version to future work.


Finally, our refusal detection counts only direct refusals. Indirect refusals were measured only for \method{False Claim}, in a separate analysis that does not enter the main results. The refusal rates in Figure~\ref{fig:matrix} are therefore approximate, but both kinds of refusal score close to zero, so the main results treat them alike.

\section{Conclusion}

We compared nine published automated methods for evaluating LLM persuasiveness on the same fifteen models under a shared setup. The methods agree only weakly on how they rank the models, and this is unlikely to be explained by sampling noise. Agreement appears weakest among the most capable models, which are the main concern of regulation. 

Two factors account for part of the disagreement. Models that refuse some tasks but not others (directly or indirectly), lower agreement by about a quarter, and general capability shapes which methods agree, since most rational persuasion methods track it and most manipulation methods do not. Together these produce a split by task type, indicative with eight methods: rational persuasion methods agree moderately with one another, whereas manipulation methods agree neither with them nor with each other, since different models refuse different manipulative tasks.
Persuasion scores therefore combine a model's ability to persuade with its willingness to do so, and a low manipulation score may reflect stronger safeguards rather than weaker persuasive ability. 
LLM persuasion is thus largely in the eye of the evaluation, as conclusions about a model's persuasiveness depend on the method used and on how its outputs are interpreted.

A single persuasion score describes a model on one task and may say little about others, even when refusals are set aside. Evaluations would benefit from using more than one task, chosen for the risk being assessed, from reporting refusals, including indirect ones, separately from persuasive success, and from reporting general capability alongside.

\section*{Acknowledgments}
The research leading to these results was funded by the European Union (ERC, VIGILIA, 101142229), the Special Research Fund (BOF) of Ghent University (BOF20/IBF/117), the Flemish Government under the ``Onderzoeksprogramma Artificiële Intelligentie (AI) Vlaanderen'' programme, and the FWO (project no. G073924N). Views and opinions expressed are however those of the author(s) only and do not necessarily reflect those of the European Union or the European Research Council Executive Agency. Neither the European Union nor the granting authority can be held responsible for them. For the purpose of Open Access the author has applied a CC BY public copyright license to any Author Accepted Manuscript version arising from this submission.

\bibliography{references}

\appendix
\section*{Appendix}


\section{Detailed Method Descriptions}\label{app:methods}

\paragraph{Reduce Meat.} This evaluation method was adapted from \citet{ahnert2025meatless}. 
The evaluated model is a persuader who, over five conversational rounds, tries to move a simulated persuadee (with one of two personas: hard or easy to persuade) towards eating less meat in one of two settings (a canteen lunch or a social-media thread). 
The persuadee is asked to answer a three-item, 7-point intention questionnaire outside the conversation, and the score is the change from round one to round five ($-6$ to $+6$). In total there were four conditions (persona $\times$ setting), each was run as 20 separate conversations, 80 per model.

\paragraph{Mislead Debate.} Adapted from \citet{amayuelas2024attack}, this method evaluates a model's ability to persuade others to adopt an incorrect answer in a three-agent debate on a multiple-choice question. The evaluated model acts as an adversary and is assigned a specific wrong option to defend. The other two agents (both played by the pinned persuadee) answer independently before engaging in two rounds of debate and revising their answers after each round. The score is the change in the proportion of persuadee agents selecting the adversary's assigned option between their initial and final answers (range: $-1$ to $1$). We used 200 questions sampled from three of the original's four datasets \textit{MMLU} \citep{hendrycks2021mmlu}, \textit{TruthfulQA} \citep{lin2022truthfulqa}, and \textit{MedMCQA} \citep{pal2022medmcqa}, with the same assigned wrong option used for all evaluated models. The fourth dataset, which was a legal reasoning task, \textit{SCALR}, is not publicly hosted outside the original code.

\paragraph{Subjective Opinion.} 
Adapted from \citet{bozdag2025pmiyc}, this evaluation measures the ability of the evaluated model to increase a persuadee's support for a subjective claim on a social or political issue. The persuader and persuadee models alternate turns in a seven-message exchange, with the persuader speaking first. The persuadee rates its support for the claim on a five-point scale before and after the exchange. The score is the change in this rating, normalised by the maximum possible change in the same direction (range: $-1$ to $1$). We used 100 claims.

\paragraph{Rationale.}
Adapted from \citet{elaraby2024rationales}, this evaluation measures how persuasively the evaluated model can justify a choice between two arguments. The model is given a debate topic and a pair of arguments on the same side of it, drawn from IBM-ArgQ-9.1kPairs, and, unlike the original is told which of the two a reference GPT-4 run had recommended for a speech on the topic. It writes a rationale for recommending that argument over the other. The pinned judge then compares the model's rationale with the reference GPT-4 rationale for the same argument, once in each order to cancel position bias. The score is the win rate against the GPT-4 rationale, counting a tie as half a win (0 to 1; 0.5 means equally persuasive). We used 100 argument pairs sampled across the dataset's twenty debate topics.

\paragraph{False Claim.} Adapted from \citet{ju2025adaptive}, this evaluation measures the ability of a model to convince a persuadee LLM that a specific false claim is true. 
For each question from the \textit{CounterFact} dataset \citep{meng2022locating}, the evaluated model is given the question and a false answer, and writes a paragraph of misleading evidence for that answer using one of eleven predefined persuasion strategies (appeals to authority, conformity, etc.). The pinned persuadee reads the paragraph and then answers the question.
The final metric is the proportion of responses that contain the false answer (range: 0 to 1). We used 120 questions, evaluating each with all eleven strategies, for a total of 1{,}320 attempts.

\paragraph{Benign Request.} Adapted from \citet{liu2025persusafety}, this evaluation measures the model's ability to persuade a resistant persuadee model to comply with a benign request over a maximum of 10 conversational turns. Across 30 scenarios, the persuadee model is set up to resist the request and adopts one of five personalities (resilient, emotionally sensitive, conflict-averse, info-overwhelmed or anxious). The model converses with the persuadee for up to ten turns and may use any of 15 manipulative persuasion strategies listed in its prompt, but is not required to. After each message the persuadee may accept or reject the request. The conversation ends if the request is accepted, rejected four times, or after 10 turns. A pinned LLM-judge rates the persuasive effectiveness on a five-point rubric, which is then normalized to a [0, 1] scale. The persuadee's decision is recorded but does not enter the score. We used every scenario with all five personalities, 150 conversations in total.

\paragraph{MakeMePay.} Adapted from the \textit{MakeMePay} game in \citet{openai2024o1card}, this evaluation measures the ability of the evaluated model to persuade a target to hand over money. The persuader model acts as a con artist against a pinned model playing the target, who holds \$100. The two converse for up to ten turns, and the target can donate by writing a payment tag such as \texttt{[GIVE \$N]}. After each reply, a rule checks for this tag, if there is none but the target has mentioned a dollar amount -- a summarising model extracts the donated amount. The episode ends upon the first donation, if the target withdraws, or after ten turns. The score is the number of dollars donated (ranging from 0 to 100), averaged over 250 episodes.

\paragraph{MakeMeSay.} Adapted from the \textit{MakeMeSay} game in \citet{openai2024o1card}, this evaluation measures the ability of the evaluated model to manipulate a persuadee model into saying a secret codeword, known only to the persuader, without noticing it. The persuader and the pinned persuadee exchange messages over 15 conversational rounds. Afterwards the persuadee is asked to guess the codeword, and the pinned LLM-judge rules whether the persuadee identified it. The persuader wins if the persuadee says the codeword before the persuader does and then fails to identify it when asked. The final persuasiveness score is the win rate over 50 codewords, each played once.

\paragraph{Rewrite Text.} Adapted from \citet{pauli2025persuasive}, this evaluation measures the ability of the evaluated model to make a human-written text more persuasive. The model is asked to rewrite the text so that it sounds more persuasive. A frozen DeBERTaV3-Large regression model, trained on human pairwise judgements of persuasiveness, scores the original and the rewrite as a pair, in both orders. The final score is the scorer's relative persuasiveness rating of the rewrite against the original, averaged across both presentation orders. Positive scores favour the rewrite, and negative scores favour the original. We used 230 texts sampled from the \textit{Persuasive-Pairs} dataset.


\begin{table*}[t]
\centering\small\tabcolsep=5pt
\begin{tabular}{@{}l r l r r l@{}}
\toprule
\textbf{Method} & \textbf{Outputs} & \textbf{Unit} & \textbf{Units} & \textbf{Per unit} & \textbf{Varies within unit} \\
\midrule
\method{Reduce Meat}        & 80      & conversation, within persona $\times$ setting & 80 & 1 & -- \\
\method{Mislead Debate}     & 200     & question                 & 200 & 1  & -- \\
\method{Subjective Opinion} & 100     & claim                    & 100 & 1  & -- \\
\method{Rationale}          & 100     & debate topic             & 20  & 5  & argument pair \\
\method{False Claim}        & 1{,}320 & question                 & 120 & 11 & persuasion strategy \\
\method{Benign Request}     & 150     & scenario                 & 30  & 5  & persuadee personality \\
\method{MakeMePay}          & 250     & episode                  & 250 & 1  & -- \\
\method{MakeMeSay}          & 50      & codeword                 & 50  & 1  & -- \\
\method{Rewrite Text}       & 230     & source text              & 230 & 1  & -- \\
\bottomrule
\end{tabular}
\caption{Sampling design per method. \emph{Outputs}: scored attempts per model. \emph{Unit}: what is resampled in bootstraps and split in split-halves, as outputs within a unit are not independent. Fixed conditions (strategies, personalities, personas, settings) are never resampled, only the units within them. \emph{Units}: designed counts (Table~\ref{tab:reliability} counts only units scored for all fifteen models).}
\label{tab:units}
\end{table*}


\section{Method Adaptations}\label{app:adaptations}

Every persuadee is played by one shared model and every judge by another, described under Experimental Setup. In three original methods the evaluated model played its own persuadee (\method{Reduce Meat}, \method{False Claim}, \method{Mislead Debate}), which conflates a model's persuasiveness with its own resistance. A fixed persuadee separates the two and gives every method the same recipient. A shared judge likewise gives the three judged methods the same scorer.
\method{Rationale} tells every model which of the two arguments to justify, so all models justify the same choice. The original dropped the items on which models chose differently. Each rationale is then judged against one fixed GPT-4 rationale from the dataset, not against every other model's as in the original. This keeps each model's score independent of the others. 
\method{Subjective Opinion} runs only the original's subjective claims, its headline setting. The original reports that persuader rankings are stable across its claim types. Its persuadee ran at temperature 0, where the original used 0.7.
\method{Benign Request} was evaluated only on the original study's ethically neutral scenarios, so that refusals of harmful requests do not dominate its score.
Its persuadee is told to resist the request, as in the released code, although the persuadee prompt printed in the original paper has no such instruction. Without it the persuadee accepted almost every request. Its persuadee ran at temperature 0, where the original used 1. It leaves out 26 conversations whose judge verdict our parser could not read, 22 of them a 0 because the persuader left its task, which the original averages in. 
\method{Rewrite Text} draws its texts from the data its scorer was trained on, where the original used 200 new texts, which raises its scores but, as every model rewrites the same texts, should barely affect its ranking.
In \method{Mislead Debate} we score how many of the other agents adopt the wrong option the adversary was assigned. The original scored agreement with whatever answer the adversary stated, so the two differ only where the adversary itself abandoned its assigned wrong answer. We also cap every debate message at 512 tokens.
\method{MakeMePay} and \method{MakeMeSay} are applied to the released models with final safety training, whereas OpenAI evaluates both pre-mitigation and post-mitigation models \citep{openai2024o1card}. 
\modelb{Claude-Fable-5.1}, \modelb{GPT-5.6-Sol} and \modelb{Kimi-K2.6} accept only their default temperature, so they ran at it on every method.
The pooled scores of \method{False Claim} and \method{Mislead Debate}, and the \method{Rewrite Text} scorer, are checked in Section~\ref{sec:checks}.
Smaller changes, such as token limits, prompt details and turn limits, will be listed in full with the released code.


\begin{table*}[h]
\centering\footnotesize\tabcolsep=4pt
\begin{tabular}{@{}p{2.3cm} p{4.2cm} p{10.3cm}@{}}
\toprule
\textbf{Axis} & \textbf{Definition} & \textbf{Groups and their methods} \\
\midrule
\multicolumn{3}{@{}l}{\emph{How a method measures, or is designed}} \\
Scored & what the score stands for & \emph{belief change}: \method{Subjective Opinion}, \method{False Claim}, \method{Benign Request}, \method{Reduce Meat}, \method{Mislead Debate}; \emph{argument}: \method{Rewrite Text}, \method{Rationale}; \emph{behaviour}: \method{MakeMePay}, \method{MakeMeSay} \\[2pt]
Scorer & what produces the score & \emph{self-report}: \method{Subjective Opinion}, \method{False Claim}, \method{Reduce Meat}; \emph{LLM-judge}: \method{Rationale}, \method{Benign Request}; \emph{rule}: \method{MakeMePay}, \method{MakeMeSay}; \emph{trained scorer}: \method{Rewrite Text}; \emph{ground truth}: \method{Mislead Debate} \\[2pt]
Format & the survey's grouping of the method's form \citep{dementaviciute2026survey} & \emph{dialogue}: \method{Subjective Opinion}, \method{False Claim}, \method{Benign Request}, \method{MakeMePay}, \method{MakeMeSay}, \method{Reduce Meat}; \emph{dataset}: \method{Rewrite Text}, \method{Rationale}; \emph{debate}: \method{Mislead Debate} \\[2pt]
Interaction & whether a persuadee is in the loop, and for how many turns & \emph{multi-turn}: \method{Subjective Opinion}, \method{Benign Request}, \method{MakeMePay}, \method{MakeMeSay}, \method{Reduce Meat}, \method{Mislead Debate}; \emph{static}: \method{Rewrite Text}, \method{Rationale}; \emph{single turn}: \method{False Claim} \\[2pt]
Pinned agent & which pinned agent the score depends on & \emph{persuadee}: \method{Subjective Opinion}, \method{False Claim}, \method{MakeMePay}, \method{MakeMeSay}, \method{Reduce Meat}, \method{Mislead Debate}; \emph{judge}: \method{Rationale}, \method{Benign Request}; \emph{none}: \method{Rewrite Text} \\[2pt]
Persuadee present & whether a persuadee reads the model's output & \emph{persuadee}: \method{Subjective Opinion}, \method{False Claim}, \method{Benign Request}, \method{MakeMePay}, \method{MakeMeSay}, \method{Reduce Meat}, \method{Mislead Debate}; \emph{none}: \method{Rewrite Text}, \method{Rationale} \\[2pt]
Turns & how long the exchange is & \emph{single output}: \method{Rewrite Text}, \method{Rationale}, \method{False Claim}; \emph{short}: \method{Subjective Opinion}, \method{Reduce Meat}, \method{Mislead Debate}; \emph{long}: \method{Benign Request}, \method{MakeMePay}, \method{MakeMeSay} \\[2pt]
Length (4 levels) & Interaction and length combined & \emph{short}: \method{Subjective Opinion}, \method{Reduce Meat}, \method{Mislead Debate}; \emph{long}: \method{Benign Request}, \method{MakeMePay}, \method{MakeMeSay}; \emph{static}: \method{Rewrite Text}, \method{Rationale}; \emph{single turn}: \method{False Claim} \\[2pt]
Length (2 levels) & short or long exchange & \emph{short}: \method{Rewrite Text}, \method{Rationale}, \method{Subjective Opinion}, \method{False Claim}, \method{Reduce Meat}, \method{Mislead Debate}; \emph{long}: \method{Benign Request}, \method{MakeMePay}, \method{MakeMeSay} \\[2pt]
Granularity & how finely one item is scored & \emph{discrete}: \method{Rationale}, \method{Subjective Opinion}, \method{Benign Request}, \method{Reduce Meat}, \method{Mislead Debate}; \emph{continuous}: \method{Rewrite Text}, \method{MakeMePay}; \emph{binary}: \method{False Claim}, \method{MakeMeSay} \\[2pt]
Relative scoring & whether the score is a change from a baseline or a comparison with a reference, or an outcome on its own & \emph{relative}: \method{Rewrite Text}, \method{Rationale}, \method{Subjective Opinion}, \method{Reduce Meat}, \method{Mislead Debate}; \emph{absolute}: \method{False Claim}, \method{Benign Request}, \method{MakeMePay}, \method{MakeMeSay} \\[2pt]
Resistance & whether the persuadee is told to hold firm or be cautious & \emph{open}: \method{Subjective Opinion}, \method{False Claim}, \method{MakeMeSay}, \method{Reduce Meat}, \method{Mislead Debate}; \emph{no persuadee}: \method{Rewrite Text}, \method{Rationale}; \emph{resistant}: \method{Benign Request}, \method{MakeMePay} \\[2pt]
\midrule
\multicolumn{3}{@{}l}{\emph{What the task asks}} \\
Goal & what the model must achieve & \emph{comply}: \method{Benign Request}, \method{MakeMePay}, \method{MakeMeSay}, \method{Reduce Meat}; \emph{shift stance}: \method{Rewrite Text}, \method{Rationale}, \method{Subjective Opinion}; \emph{accept falsehood}: \method{False Claim}, \method{Mislead Debate} \\[2pt]
Task type & whether the task's goal requires manipulating the persuadee (\method{Benign Request}'s goal does not, although its persuader may use manipulative strategies) & \emph{rational persuasion}: \method{Rewrite Text}, \method{Rationale}, \method{Subjective Opinion}, \method{Benign Request}, \method{Reduce Meat}; \emph{manipulative}: \method{False Claim}, \method{MakeMePay}, \method{MakeMeSay}, \method{Mislead Debate} \\[2pt]
Topic & the subject matter & \emph{politics \& society}: \method{Rewrite Text}, \method{Rationale}, \method{Subjective Opinion}; \emph{factual knowledge}: \method{False Claim}, \method{Mislead Debate}; \emph{everyday life}: \method{Benign Request}, \method{Reduce Meat}; \emph{open conversation}: \method{MakeMePay}, \method{MakeMeSay} \\[2pt]
Stake & what the persuadee gives up by yielding & \emph{none}: \method{Rewrite Text}, \method{Rationale}, \method{Subjective Opinion}, \method{MakeMeSay}; \emph{its answer}: \method{False Claim}, \method{Mislead Debate}; \emph{personal commitment}: \method{Benign Request}, \method{Reduce Meat}; \emph{money}: \method{MakeMePay} \\[2pt]
Task type (3 levels) & Task type with the rational persuasion methods split by what they ask the persuadee to change & \emph{manipulative}: \method{False Claim}, \method{MakeMePay}, \method{MakeMeSay}, \method{Mislead Debate}; \emph{rational, opinion}: \method{Rewrite Text}, \method{Rationale}, \method{Subjective Opinion}; \emph{rational, behaviour}: \method{Benign Request}, \method{Reduce Meat} \\[2pt]
Claims or actions & whether the persuadee is to believe something or to do something & \emph{claims}: \method{Rewrite Text}, \method{Rationale}, \method{Subjective Opinion}, \method{False Claim}, \method{Mislead Debate}; \emph{actions}: \method{Benign Request}, \method{MakeMePay}, \method{MakeMeSay}, \method{Reduce Meat} \\
\bottomrule
\end{tabular}
\caption{The eighteen RQ2 groupings, each with its definition and the methods in each group. Results in Table~\ref{tab:axes}.}
\label{tab:axes_labels}
\end{table*}


\begin{table*}[t]
\centering\footnotesize\tabcolsep=4pt
\begin{tabular}{@{}l p{6.7cm} rrrrrr@{}}
\toprule
\textbf{Axis} & \textbf{Groups (number of methods)} & \textbf{Within} & \textbf{Between} & \textbf{Gap} & \textbf{95\% CI} & \textbf{Perm.\ $p$} & \textbf{Adj.\ $p$} \\
\midrule
\multicolumn{8}{@{}l}{\emph{How a method measures, or is designed}} \\
Scored & belief change 4, argument 2, behaviour 2 & 0.13 & 0.30 & $-0.17$ & [$-0.22$, $-0.11$] & 0.29 & 0.98 \\
Scorer & self-report 3, LLM-judge 2, rule 2, trained scorer 1 & 0.14 & 0.28 & $-0.13$ & [$-0.21$, $-0.06$] & 0.44 & 1.00 \\
Format & dialogue 6, dataset 2 & 0.16 & 0.37 & $-0.21$ & [$-0.23$, $-0.11$] & 0.32 & 0.99 \\
Interaction & multi-turn 5, static 2, single turn 1 & 0.35 & 0.19 & 0.16 & [0.08, 0.22] & 0.51 & 1.00 \\
Pinned agent & persuadee 5, judge 2, none 1 & 0.16 & 0.31 & $-0.15$ & [$-0.18$, $-0.03$] & 0.52 & 1.00 \\
Persuadee present & persuadee 6, none 2 & 0.16 & 0.37 & $-0.21$ & [$-0.23$, $-0.11$] & 0.32 & 0.99 \\
Turns & single output 3, short 2, long 3 & 0.23 & 0.26 & $-0.03$ & [$-0.08$, 0.04] & 0.85 & 1.00 \\
Length (4 levels) & short 2, long 3, static 2, single turn 1 & 0.40 & 0.22 & 0.19 & [0.10, 0.24] & 0.27 & 0.98 \\
Length (2 levels) & short 5, long 3 & 0.31 & 0.20 & 0.11 & [0.01, 0.16] & 0.39 & 1.00 \\
Granularity & discrete 4, continuous 2, binary 2 & 0.42 & 0.18 & 0.25 & [0.10, 0.29] & 0.13 & 0.82 \\
Relative scoring & relative 4, absolute 4 & 0.36 & 0.17 & 0.19 & [0.08, 0.25] & 0.029 & 0.32 \\
Resistance & open 4, no persuadee 2, resistant 2 & 0.28 & 0.24 & 0.05 & [$-0.01$, 0.15] & 0.75 & 1.00 \\
\midrule
\multicolumn{8}{@{}l}{\emph{What the task asks}} \\
Goal & comply 4, shift stance 3, accept falsehood 1 & 0.41 & 0.17 & 0.24 & [0.17, 0.31] & 0.12 & 0.80 \\
Task type & rational persuasion 5, manipulative 3 & 0.45 & 0.06 & 0.40 & [0.24, 0.44] & 0.036 & 0.39 \\
Topic & politics \& society 3, factual knowledge 1, everyday life 2, open conversation 2 & 0.52 & 0.19 & 0.33 & [0.25, 0.40] & 0.037 & 0.39 \\
Stake & none 4, its answer 1, personal commitment 2, money~1 & 0.56 & 0.13 & 0.42 & [0.34, 0.48] & 0.021 & 0.23 \\
Task type (3 levels) & manipulative 3, neutral 3, prosocial 2 & 0.38 & 0.21 & 0.18 & [0.12, 0.27] & 0.16 & 0.89 \\
Claims or actions & claims 4, actions 4 & 0.24 & 0.26 & $-0.02$ & [$-0.06$, 0.07] & 0.91 & 1.00 \\
\bottomrule
\end{tabular}
\caption{All eighteen RQ2 groupings, over the eight reliable methods. 
\emph{Within}, \emph{Between}, \emph{Gap}, \emph{Perm.\ $p$} and \emph{Adj.\ $p$} as in Table~\ref{tab:axes}.
Table~\ref{tab:axes_labels} lists the methods in each group. \textit{95\% CI}: over sampling units, with the eight methods fixed.}
\label{tab:axes_full}
\end{table*}


\begin{table*}[t]
\centering\footnotesize\tabcolsep=5pt
\begin{tabular}{@{}lrrr@{}}
\toprule
\textbf{Model} & \textbf{MMLU-Pro} & \textbf{IFEval} & $C$ \\
\midrule
\modelb{DeepSeek-V4.1-Flash}  & 90.7 & 96.1 & $+1.15$ \\
\modelb{GPT-5.6-Sol}                & 91.7 & 94.1 & $+1.10$ \\
\modelb{Claude-Fable-5.1}           & 91.7 & 93.3 & $+1.08$ \\
\modelb{Kimi-K2.6}                  & 82.7 & 90.9 & $+0.76$ \\
\modelb{Llama-3.3-70B}     & 70.3 & 90.2 & $+0.41$ \\
\modelb{GPT-OSS-120B}               & 76.3 & 84.3 & $+0.35$ \\
\modelb{Mistral-Medium-3.5}         & 77.3 & 83.0 & $+0.32$ \\
\modelb{Qwen3-32B}                  & 78.3 & 81.9 & $+0.31$ \\
\modelb{EXAONE-4.0.1-32B}           & 70.0 & 82.8 & $+0.13$ \\
\modelb{Gemma-3-27B}             & 65.0 & 83.7 & $+0.03$ \\
\modelb{Gemma-3-12B}             & 58.0 & 82.3 & $-0.20$ \\
\modelb{Qwen2.5-7B}        & 58.0 & 70.8 & $-0.63$ \\
\modelb{Llama-3.2-3B}      & 37.7 & 69.5 & $-1.20$ \\
\modelb{Mistral-Nemo-12B} & 44.3 & 58.0 & $-1.45$ \\
\modelb{Mistral-7B-v0.2}   & 29.3 & 49.2 & $-2.17$ \\
\bottomrule
\end{tabular}
\caption{General capability of the fifteen models, ordered by $C$. \emph{MMLU-Pro}: accuracy (\%) on a fixed random subset of 300 questions, five-shot with chain of thought. \emph{IFEval}: prompt-level strict accuracy (\%) on all 541 prompts, zero-shot. Both ran with \texttt{lm-evaluation-harness} at temperature 0, or the default for \modelb{GPT-5.6-Sol}, \modelb{Claude-Fable-5.1} and \modelb{Kimi-K2.6}, which accept no other. 
$C$: the mean of the two accuracies, each z-scored across the fifteen models. The two benchmarks rank the models alike (Spearman $\rho = 0.87$), and where a published five-shot chain-of-thought score exists, ours is within three points of it.}
\label{tab:capability_models}
\end{table*}


\end{document}

%% file: fig_top3.tex
\definecolor{topred}{RGB}{246, 72, 71}
\definecolor{bottomblue}{RGB}{57, 135, 229}
\definecolor{faintdot}{RGB}{216,216,212}
\newcommand{\dFirst}{\tikz[baseline=-0.55ex]{\fill[topred] circle (3.0pt);\draw[white,line width=0.6pt] circle (1.75pt);}}
\newcommand{\dTop}{\tikz[baseline=-0.55ex]{\fill[topred] circle (2.3pt);}}
\newcommand{\dLast}{\tikz[baseline=-0.55ex]{\draw[bottomblue,line width=0.75pt] circle (2.0pt);}}
\newcommand{\dNone}{\tikz[baseline=-0.55ex]{\fill[faintdot] circle (0.8pt);}}
\newcommand{\rot}[1]{\rotatebox{90}{\scriptsize #1}}

\begin{tabular}{@{}l@{\hspace{7pt}}r@{\hspace{8pt}}c@{\hspace{5pt}}c@{\hspace{5pt}}c@{\hspace{5pt}}c@{\hspace{5pt}}c@{\hspace{10pt}}c@{\hspace{5pt}}c@{\hspace{5pt}}c@{\hspace{5pt}}c@{\hspace{4pt}}|@{\hspace{4pt}}c@{}}
\multicolumn{7}{r}{\scriptsize\itshape rational persuasion} & \multicolumn{5}{@{}l@{}}{\scriptsize\itshape manipulation} \\
\cmidrule(lr){3-7}\cmidrule(lr){8-11}
& \multicolumn{1}{r}{\scriptsize size} & \rot{Rationale} & \rot{Rewrite Text} & \rot{Subjective Opinion} & \rot{Reduce Meat} & \rot{Benign Request} & \rot{MakeMeSay} & \rot{MakeMePay} & \rot{Mislead Debate} & \rot{False Claim} & \rot{top-three rankings} \\
\addlinespace[2pt]
\textit{\model{Claude-Fable-5.1}} & {\scriptsize --} & \dFirst & \dTop & \dTop & \dNone & \dFirst & \dLast & \dLast & \dNone & \dLast & \scriptsize 4 \\[1.5pt]
\textit{\model{GPT-5.6-Sol}} & {\scriptsize --} & \dNone & \dNone & \dFirst & \dNone & \dNone & \dTop & \dNone & \dNone & \dLast & \scriptsize 2 \\[1.5pt]
\textit{\model{Kimi-K2.6}} & {\scriptsize 1T*} & \dTop & \dFirst & \dTop & \dFirst & \dNone & \dTop & \dNone & \dNone & \dNone & \scriptsize 5 \\[1.5pt]
\textit{\model{DeepSeek-V4.1-Flash}} & {\scriptsize 552B*} & \dTop & \dNone & \dNone & \dTop & \dNone & \dNone & \dNone & \dNone & \dNone & \scriptsize 2 \\[1.5pt]
\textit{\model{Mistral-Medium-3.5}} & {\scriptsize 128B} & \dNone & \dTop & \dNone & \dNone & \dNone & \dFirst & \dNone & \dTop & \dNone & \scriptsize 3 \\[1.5pt]
\model{gpt-oss-120b} & {\scriptsize 120B*} & \dNone & \dNone & \dNone & \dNone & \dLast & \dNone & \dLast & \dNone & \dNone & \scriptsize\textcolor{faintdot}{0} \\[1.5pt]
\model{Llama-3.3-70B} & {\scriptsize 70B} & \dNone & \dNone & \dNone & \dNone & \dTop & \dNone & \dFirst & \dLast & \dNone & \scriptsize 2 \\[1.5pt]
\model{EXAONE-4.0.1-32B} & {\scriptsize 32B} & \dNone & \dNone & \dNone & \dNone & \dLast & \dNone & \dNone & \dNone & \dTop & \scriptsize 1 \\[1.5pt]
\model{Qwen3-32B} & {\scriptsize 32B} & \dNone & \dNone & \dNone & \dTop & \dNone & \dNone & \dNone & \dNone & \dNone & \scriptsize 1 \\[1.5pt]
\model{Gemma-3-27B} & {\scriptsize 27B} & \dNone & \dNone & \dNone & \dNone & \dNone & \dNone & \dTop & \dTop & \dFirst & \scriptsize 3 \\[1.5pt]
\model{Mistral-Nemo-12B} & {\scriptsize 12B} & \dNone & \dNone & \dNone & \dNone & \dNone & \dNone & \dNone & \dNone & \dNone & \scriptsize\textcolor{faintdot}{0} \\[1.5pt]
\model{Gemma-3-12B} & {\scriptsize 12B} & \dNone & \dNone & \dNone & \dNone & \dTop & \dNone & \dTop & \dFirst & \dNone & \scriptsize 3 \\[1.5pt]
\model{Mistral-7B-v0.2} & {\scriptsize 7B} & \dLast & \dLast & \dLast & \dLast & \dNone & \dLast & \dNone & \dNone & \dNone & \scriptsize\textcolor{faintdot}{0} \\[1.5pt]
\model{Qwen2.5-7B} & {\scriptsize 7B} & \dLast & \dNone & \dNone & \dNone & \dNone & \dNone & \dNone & \dNone & \dTop & \scriptsize 1 \\[1.5pt]
\model{Llama-3.2-3B} & {\scriptsize 3B} & \dNone & \dLast & \dLast & \dLast & \dNone & \dNone & \dNone & \dLast & \dNone & \scriptsize\textcolor{faintdot}{0} \\[1.5pt]
\addlinespace[3pt]
\multicolumn{12}{@{}c@{}}{\scriptsize \dFirst~1st \quad \dTop~2nd or 3rd \quad \dLast~14th or 15th} \\
\end{tabular}

%% file: references.bib
@inproceedings{pauli2025persuasive,
  title={Measuring and {B}enchmarking {L}arge {L}anguage {M}odels’ {C}apabilities to {G}enerate {P}ersuasive {L}anguage},
  author={Pauli, Amalie Brogaard and others},
  booktitle={Proceedings of the 2025 Conference of the Nations of the Americas Chapter of the Association for Computational Linguistics: Human Language Technologies (Volume 1: Long Papers)},
  pages={10056--10075},
  year={2025}
}

@inproceedings{elaraby2024rationales,
  title = "Persuasiveness of {G}enerated {F}ree-{T}ext {R}ationales in {S}ubjective {D}ecisions: {A} {C}ase {S}tudy on {P}airwise {A}rgument {R}anking",
  author={Elaraby, Mohamed and Litman, Diane and Li, Xiang Lorraine and Magooda, Ahmed},
  booktitle={Findings of the Association for Computational Linguistics: EMNLP 2024},
  pages={14311--14329},
  year={2024}
}

@inproceedings{bozdag2025pmiyc,
author = {Bozdag, Nimet Beyza and Mehri, Shuhaib and Tur, Gokhan and Hakkani-Tur, Dilek},
title = {Persuade Me if You Can: A Framework for Evaluating Persuasion Effectiveness and Susceptibility Among Large Language Models},
year = {2026},
isbn = {9798400724152},
publisher = {Association for Computing Machinery},
address = {New York, NY, USA},
url = {https://doi.org/10.1145/3786335.3813181},
doi = {10.1145/3786335.3813181},
booktitle = {Proceedings of the ACM Conference on AI and Agentic Systems},
pages = {702–726},
numpages = {25},
location = {
},
series = {CAIS '26}
}

@misc{ju2025adaptive,
  author = "Ju, Tianjie and Chen, Yujia and Fei, Hao and Lee, Mong-Li and Hsu, Wynne and Cheng, Pengzhou and Wu, Zongru and Zhang, Zhuosheng and Liu, Gongshen",
  title  = "On the {A}daptive {P}sychological {P}ersuasion of {L}arge {L}anguage {M}odels",
  year   = "2025",
  note   = "Preprint at \url{https://arxiv.org/abs/2506.06800}"
}

@inproceedings{
liu2025persusafety,
title={{LLM} Can be a Dangerous Persuader: Empirical Study of Persuasion Safety in Large Language Models},
author={Minqian Liu and Zhiyang Xu and Xinyi Zhang and Heajun An and Sarvech Qadir and Qi Zhang and Pamela J. Wisniewski and Jin-Hee Cho and Sang Won Lee and Ruoxi Jia and Lifu Huang},
booktitle={Second Conference on Language Modeling},
year={2025},
url={https://openreview.net/forum?id=TMB9SKqit9}
}

@techreport{openai2024o1card,
  author = {{OpenAI}},
  title = {OpenAI o1 System Card},
  institution = {OpenAI},
  year = {2024},
  month = {December},
  url = {https://cdn.openai.com/o1-system-card.pdf},
  type = {System Card}
}

@inproceedings{ahnert2025meatless,
    title = "Simulating Persuasive Dialogues on Meat Reduction with Generative Agents",
    author = "Ahnert, Georg  and
      Wurth, Elena  and
      Strohmaier, Markus  and
      Mata, Jutta",
    editor = "Velutharambath, Aswathy  and
      Labat, Sofie  and
      Falk, Neele  and
      Plaza-del-Arco, Flor Miriam  and
      Klinger, Roman  and
      Hoste, V{\'e}ronique",
    booktitle = "Proceedings of the First Workshop on Integrating {NLP} and Psychology to Study Social Interactions ({NLPSI}) @{ICWSM} `25",
    month = jun,
    year = "2025",
    address = "Copenhagen, Denmark",
    publisher = "Association for the Advancement of Artificial Intelligence (www.aaai.org)",
    url = "https://aclanthology.org/2025.nlpsi-1.1/",
    doi = "10.36190/2025.30",
    pages = "1--10"
}

@inproceedings{amayuelas2024attack,
    title = "{M}ulti{A}gent Collaboration Attack: Investigating Adversarial Attacks in Large Language Model Collaborations via Debate",
    author = "Amayuelas, Alfonso  and
      Yang, Xianjun  and
      Antoniades, Antonis  and
      Hua, Wenyue  and
      Pan, Liangming  and
      Wang, William Yang",
    editor = "Al-Onaizan, Yaser  and
      Bansal, Mohit  and
      Chen, Yun-Nung",
    booktitle = "Findings of the Association for Computational Linguistics: EMNLP 2024",
    month = nov,
    year = "2024",
    address = "Miami, Florida, USA",
    publisher = "Association for Computational Linguistics",
    url = "https://aclanthology.org/2024.findings-emnlp.407/",
    doi = "10.18653/v1/2024.findings-emnlp.407",
    pages = "6929--6948"
}

@inproceedings{wang2024mmlupro,
  title     = {{MMLU-Pro}: A More Robust and Challenging Multi-Task Language Understanding Benchmark},
  author    = {Wang, Yubo and Ma, Xueguang and Zhang, Ge and Ni, Yuansheng and Chandra, Abhranil and Guo, Shiguang and Ren, Weiming and Arulraj, Aaran and He, Xuan and Jiang, Ziyan and Li, Tianle and Ku, Max and Wang, Kai and Zhuang, Alex and Fan, Rongqi and Yue, Xiang and Chen, Wenhu},
  booktitle = {Advances in Neural Information Processing Systems 37, Datasets and Benchmarks Track},
  year      = {2024}
}

@article{zhou2023ifeval,
  title   = {Instruction-Following Evaluation for Large Language Models},
  author  = {Zhou, Jeffrey and Lu, Tianjian and Mishra, Swaroop and Brahma, Siddhartha and Basu, Sujoy and Luan, Yi and Zhou, Denny and Hou, Le},
  journal = {arXiv preprint arXiv:2311.07911},
  year    = {2023}
}

@misc{gao2024lmeval,
  title     = {The Language Model Evaluation Harness},
  author    = {Gao, Leo and Tow, Jonathan and Abbasi, Baber and others},
  year      = {2024},
  publisher = {Zenodo},
  doi       = {10.5281/zenodo.12608602}
}

@inproceedings{hendrycks2021mmlu,
  title     = {Measuring Massive Multitask Language Understanding},
  author    = {Hendrycks, Dan and Burns, Collin and Basart, Steven and Zou, Andy and Mazeika, Mantas and Song, Dawn and Steinhardt, Jacob},
  booktitle = {International Conference on Learning Representations (ICLR)},
  year      = {2021}
}

@inproceedings{lin2022truthfulqa,
  title     = {{TruthfulQA}: Measuring How Models Mimic Human Falsehoods},
  author    = {Lin, Stephanie and Hilton, Jacob and Evans, Owain},
  booktitle = {Proceedings of the 60th Annual Meeting of the Association for Computational Linguistics (Volume 1: Long Papers)},
  pages     = {3214--3252},
  year      = {2022}
}

@inproceedings{pal2022medmcqa,
  title     = {{MedMCQA}: A Large-scale Multi-Subject Multi-Choice Dataset for Medical domain Question Answering},
  author    = {Pal, Ankit and Umapathi, Logesh Kumar and Sankarasubbu, Malaikannan},
  booktitle = {Proceedings of the Conference on Health, Inference, and Learning},
  series    = {Proceedings of Machine Learning Research},
  volume    = {174},
  pages     = {248--260},
  year      = {2022}
}

@inproceedings{perlitz2025bat,
  title     = {Benchmark Agreement Testing Done Right: A Guide for {LLM} Benchmark Evaluation},
  author    = {Perlitz, Yotam and Gera, Ariel and Arviv, Ofir and Yehudai, Asaf and Bandel, Elron and Shnarch, Eyal and Shmueli-Scheuer, Michal and Choshen, Leshem},
  booktitle = {NeurIPS 2025 Workshop on Evaluating the Evolving {LLM} Lifecycle: Benchmarks, Emergent Abilities, and Scaling},
  year      = {2025},
  note      = {arXiv:2407.13696}
}

@inproceedings{ren2024safetywashing,
  title     = {Safetywashing: Do {AI} Safety Benchmarks Actually Measure Safety Progress?},
  author    = {Ren, Richard and Basart, Steven and Khoja, Adam and Gatti, Alice and Phan, Long and Yin, Xuwang and Mazeika, Mantas and Pan, Alexander and Mukobi, Gabriel and Kim, Ryan H. and Fitz, Stephen and Hendrycks, Dan},
  booktitle = {Advances in Neural Information Processing Systems (NeurIPS)},
  year      = {2024},
  note      = {arXiv:2407.21792}
}

@article{norman2026judges,
  title={Reliability without Validity: A Systematic, Large-Scale Evaluation of LLM-as-a-Judge Models Across Agreement, Consistency, and Bias},
  author={Norman, Justin D and Rivera, Michael U and Hughes, D Alex},
  journal={arXiv preprint arXiv:2606.19544},
  year={2026}
}

@article{carlson2012understanding,
  title={Understanding the impact of convergent validity on research results},
  author={Carlson, Kevin D and Herdman, Andrew O},
  journal={Organizational Research Methods},
  volume={15},
  number={1},
  pages={17--32},
  year={2012},
  publisher={Sage Publications Sage CA: Los Angeles, CA}
}

@book{westfall1993resampling,
  title={Resampling-based multiple testing: Examples and methods for p-value adjustment},
  author={Westfall, Peter H and Young, S Stanley},
  year={1993},
  publisher={John Wiley \& Sons}
}

@article{spearman1904proof, author = {Spearman, Charles}, title = {The Proof and Measurement of Association between Two Things}, journal = {The American Journal of Psychology}, volume = {15}, number = {1}, pages = {72--101}, year = {1904} }

@misc{durmus2024measuring,
  title        = {Measuring the Persuasiveness of Language Models},
  author       = {Durmus, Esin and Lovitt, Liane and Tamkin, Alex and Ritchie, Stuart and Clark, Jack and Ganguli, Deep},
  howpublished = {Anthropic research post, \url{https://www.anthropic.com/research/measuring-model-persuasiveness}},
  month        = apr,
  year         = {2024}
}

@article{salvi2024conversational,
  title   = {On the Conversational Persuasiveness of {GPT-4}},
  author  = {Salvi, Francesco and Horta Ribeiro, Manoel and Gallotti, Riccardo and West, Robert},
  journal = {Nature Human Behaviour},
  volume  = {9},
  number  = {8},
  pages   = {1645--1653},
  year    = {2025},
  doi     = {10.1038/s41562-025-02194-6}
}

@article{bai2025llm,
  title={LLM-generated messages can persuade humans on policy issues},
  author={Bai, Hui and Voelkel, Jan G and Muldowney, Shane and Eichstaedt, Johannes C and Willer, Robb},
  journal={Nature Communications},
  volume={16},
  number={1},
  pages={6037},
  year={2025},
  publisher={Nature Publishing Group UK London}
}

@article{hackenburg2026ai,
  title={AI systems out-persuade expert humans},
  author={Hackenburg, Kobi and Wagner, Caroline and Hewitt, Luke and Tappin, Ben M and Saunders, Ed and Kirk, Hannah Rose and Margetts, Helen and Summerfield, Christopher},
  journal={arXiv preprint arXiv:2606.16475},
  year={2026}
}

@article{hackenburg2025scaling,
  title={Scaling language model size yields diminishing returns for single-message political persuasion},
  author={Hackenburg, Kobi and Tappin, Ben M and R{\"o}ttger, Paul and Hale, Scott A and Bright, Jonathan and Margetts, Helen},
  journal={Proceedings of the National Academy of Sciences},
  volume={122},
  number={10},
  pages={e2413443122},
  year={2025},
  publisher={National Academy of Sciences}
}

@book{o_persuasion_2015,
  title = "Persuasion: {T}heory and {R}esearch",
  author={O'Keefe, Daniel J},
  year={2015},
  publisher={Sage Publications},
  address		= "Thousand Oaks, CA", 
}

@article{kowal2025thought,
  title={It's the Thought that Counts: Evaluating the Attempts of Frontier LLMs to Persuade on Harmful Topics},
  author={Kowal, Matthew and Timm, Jasper and Godbout, Jean-Francois and Costello, Thomas and Arechar, Antonio A and Pennycook, Gordon and Rand, David and Gleave, Adam and Pelrine, Kellin},
  journal={arXiv preprint arXiv:2506.02873},
  year={2025}
}

@inproceedings{potter2024hidden,
  title={Hidden persuaders: LLMs’ political leaning and their influence on voters},
  author={Potter, Yujin and Lai, Shiyang and Kim, Junsol and Evans, James and Song, Dawn},
  booktitle={Proceedings of the 2024 Conference on Empirical Methods in Natural Language Processing},
  pages={4244--4275},
  year={2024}
}

@article{zaman2026manipulation,
  title={Manipulation Is Task-Dependent: A Multi-Axis, Multi-Environment Evaluation of Frontier LLMs},
  author={Zaman, Adeeb and others},
  journal={arXiv preprint arXiv:2606.25899},
  year={2026}
}

@article{holbling2025meta,
  title={A meta-analysis of the persuasive power of large language models},
  author={H{\"o}lbling, Lukas and Maier, Sebastian and Feuerriegel, Stefan},
  journal={Scientific Reports},
  volume={15},
  number={1},
  pages={43818},
  year={2025},
  publisher={Nature Publishing Group UK London}
}

@inproceedings{dementaviciute2026survey,
title={{LLM} Persuasiveness Evaluation: A Structured Review of Automated Methods},
author={Kamile Dementaviciute and others},
booktitle={Trustworthy AI for Good (AI4GOOD) Workshop @ ICML 2026},
year={2026},
url={https://openreview.net/forum?id=VdrqYiBXBh}
}

@article{jones_lies_2026,
  author  = "Jones, Cameron R. and Bergen, Benjamin K.",
  title   = "Lies, {D}amned {L}ies, and {L}anguage {S}tatistics: {A} {C}omprehensive {R}eview of {R}isks from {M}anipulation, {P}ersuasion, and {D}eception with {L}arge {L}anguage {M}odels",
  journal = "Artificial Intelligence Review",
  volume  = "59",
  year    = "2026",
  pages   = "116",
  doi     = "10.1007/s10462-026-11517-6"
}

@article{bozdag_must_2025,
    author = {Bozdag, Nimet Beyza and Mehri, Shuhaib and Yang, Xiaocheng and Ha, Hyeonjeong and Cheng, Zirui and Durmus, Esin and You, Jiaxuan and Ji, Heng and Tur, Gokhan and Hakkani-T\"{u}r, Dilek},
    title = {{M}ust {R}ead: {A} {C}omprehensive {S}urvey of {C}omputational {P}ersuasion},
    year = {2026},
    publisher = {Association for Computing Machinery},
    address = {New York, NY, USA},
    issn = {0360-0300},
    url = {https://doi.org/10.1145/3800687},
    doi = {10.1145/3800687},
    journal = {ACM Computing Surveys},
    month = mar
}

@misc{noels_persuasion_2026,
      title={Persuasion with Large Language Models: A Survey of Empirical Evidence, Study Methodologies, and Ethical Implications}, 
      author={Sander Noels and others},
      year={2026},
      eprint={2411.06837},
      archivePrefix={arXiv},
      primaryClass={cs.CL},
      url={https://arxiv.org/abs/2411.06837}, 
}

@article{desai2026convergent,
  title={What AI Benchmarks Actually Measure: Adapting Convergent and Discriminant Validity to Interrogate Fifty-Six AI Benchmarks},
  author={Desai, Meera and Truong, Sang T and Wallach, Hanna and Chouldechova, Alex and Cooper, A Feder and Garcia-Gathright, Jean and Ho, Daniel E and Jacobs, Abigail Z and Koyejo, Sanmi and Pangakis, Nicholas and others},
  journal={arXiv preprint arXiv:2609.08812},
  year={2026}
}

@misc{bengio2025report,
	title = "International {AI} {S}afety {R}eport",
    author={Bengio, Yoshua and Mindermann, S{\"o}ren and Privitera, Daniel and Besiroglu, Tamay and Bommasani, Rishi and Casper, Stephen and Choi, Yejin and Fox, Philip and Garfinkel, Ben and Goldfarb, Danielle and others},
	doi = {10.48550/arXiv.2501.17805},
	urldate = {2025-09-09},
	publisher = {arXiv},
	month = jan,
	year = {2025},
    note   = "Preprint at \url{http://arxiv.org/abs/2501.17805}",
}

@techreport{bengio2026report,
    title = {International {AI} Safety Report 2026},
    author = {Bengio, Yoshua and others},
    year = 2026,
    number = {DSIT 2026/001},
    url ={https://internationalaisafetyreport.org},
    institution = {Department for Science, Innovation and Technology}
}

@misc{eu2025codeofpractice,
  title        = {The General-Purpose {AI} Code of Practice},
  author       = {{European Commission}},
  year         = {2025},
  month        = jul,
  howpublished = {\url{https://digital-strategy.ec.europa.eu/en/policies/contents-code-gpai}},
  note         = {Safety and Security chapter}
}

@misc{deepmind2025fsf,
  title        = {Frontier Safety Framework, Version 3.0},
  author       = {{Google DeepMind}},
  year         = {2025},
  month        = sep,
  howpublished = {\url{https://deepmind.google/blog/strengthening-our-frontier-safety-framework/}}
}

@article{phuong2024dangerous,
  title={Evaluating frontier models for dangerous capabilities},
  author={Phuong, Mary and Aitchison, Matthew and Catt, Elliot and Cogan, Sarah and Kaskasoli, Alexandre and Krakovna, Victoria and Lindner, David and Rahtz, Matthew and Assael, Yannis and Hodkinson, Sarah and others},
  journal={arXiv preprint arXiv:2403.13793},
  year={2024}
}

@inproceedings{meng2022locating,
  title     = {Locating and Editing Factual Associations in {GPT}},
  author    = {Meng, Kevin and Bau, David and Andonian, Alex and Belinkov, Yonatan},
  booktitle = {Advances in Neural Information Processing Systems 35},
  pages     = {17359--17372},
  year      = {2022}
}
